\pdfoutput=1
\documentclass[11pt]{article}
\usepackage{acl}
\usepackage{times}
\usepackage{latexsym}

\usepackage[T1]{fontenc}
\usepackage[utf8]{inputenc}
\usepackage{inconsolata}
\usepackage{hhline}

\usepackage{microtype}
\usepackage{graphicx}
\usepackage{booktabs} 
\usepackage{tabularx} 

\usepackage{enumitem}
\usepackage{hyperref}
\usepackage{algorithm}
\usepackage{algpseudocode}
\usepackage{xspace}
\usepackage{float}

\newcommand{\Input}{\State \textbf{Input:} }
\newcommand{\Output}{\State \textbf{Output:} }

\usepackage{amsmath}
\usepackage{amssymb}
\usepackage{mathtools}
\usepackage{caption}

\usepackage{subcaption}
\usepackage{multirow}
\usepackage{multicol}
\usepackage{amsthm}

\definecolor{light-gray}{gray}{0.8}
\definecolor{light-gray0}{gray}{0.9}

\usepackage{listings}

\usepackage{color}
\definecolor{darkblue}{rgb}{0.0,0.0,0.6}
\definecolor{cyan}{rgb}{0.0,0.6,0.6}

\lstdefinelanguage{XML}
{
  morestring=[b]",
  morestring=[s]{>}{<},
  morecomment=[s]{<?}{?>},
  stringstyle=\color{black},
  identifierstyle=\color{darkblue},
  keywordstyle=\color{cyan},
  morekeywords={xmlns,version,type}
}

\usepackage[many]{tcolorbox}
\usepackage{fontawesome5}

\usepackage[textsize=tiny]{todonotes}

\newcommand{\ours}{FGO\xspace}

\author{
\textbf{Jiale Liu}\textsuperscript{\rm 1}, 
\textbf{Yifan Zeng}\textsuperscript{\rm 2}, 
\textbf{Shaokun Zhang}\textsuperscript{\rm 1}, 
\textbf{Chi Zhang}\textsuperscript{\rm 3}, 
\textbf{Malte Højmark-Bertelsen}\textsuperscript{\rm 4}, \\
\textbf{Marie Normann Gadeberg}\textsuperscript{\rm 4},
\textbf{Huazheng Wang}\textsuperscript{\rm 2}, 
\textbf{Qingyun Wu}\textsuperscript{\rm 1} \\
\textsuperscript{\rm 1}Pennsylvania State University\quad
\textsuperscript{\rm 2}Oregon State University\\
\textsuperscript{\rm 3}The University of Texas at Austin\quad
\textsuperscript{\rm 4}Beyond Work \\
\texttt{jiale.liu@psu.edu}
 }

\begin{document}
\title{Divide, Optimize, Merge: Fine-Grained LLM Agent Optimization at Scale}

\maketitle

\begin{abstract}
LLM-based optimization has shown remarkable potential in enhancing agentic systems. However, the conventional approach of prompting LLM optimizer with the whole training trajectories on training dataset in a single pass becomes untenable as datasets grow, leading to context window overflow and degraded pattern recognition. To address these challenges, we propose Fine-Grained Optimization (\ours), a scalable framework that divides large optimization tasks into manageable subsets, performs targeted optimizations, and systematically combines optimized components through progressive merging.
Evaluation across ALFWorld, LogisticsQA, and GAIA benchmarks demonstrate that \ours outperforms existing approaches by 1.6-8.6\% while reducing average prompt token consumption by 56.3\%. Our framework provides a practical solution for scaling up LLM-based optimization of increasingly sophisticated agent systems. Further analysis demonstrates that \ours achieves the most consistent performance gain in all training dataset sizes, showcasing its scalability and efficiency.
\end{abstract}

\section{Introduction}

Large Language Models (LLMs) have emerged as powerful optimizers for LLM systems, capable of analyzing execution trajectories and refining system modules like prompts~\citep{yang2023large,zhou2022large,khattab2023dspy,opsahl2024optimizing}, tools~\citep{qian2023creator,zhangoffline,zhang2024ecoact,wang2024trove}. These agentic systems have shown promising results in enhancing agent performance across various domains, including reasoning~\citep{cheng2024trace,zelikman2023self}, software engineering~\citep{jimenez2023swe,pan2024training}, data analysis~\citep{hu2024infiagent,jing2024dsbench}, computer using~\citep{wang2025mobile,xie2025osworld,abuelsaad2024agent,xia2025gui}.

\begin{figure}[t]
    \centering
    \includegraphics[width=.95\linewidth]{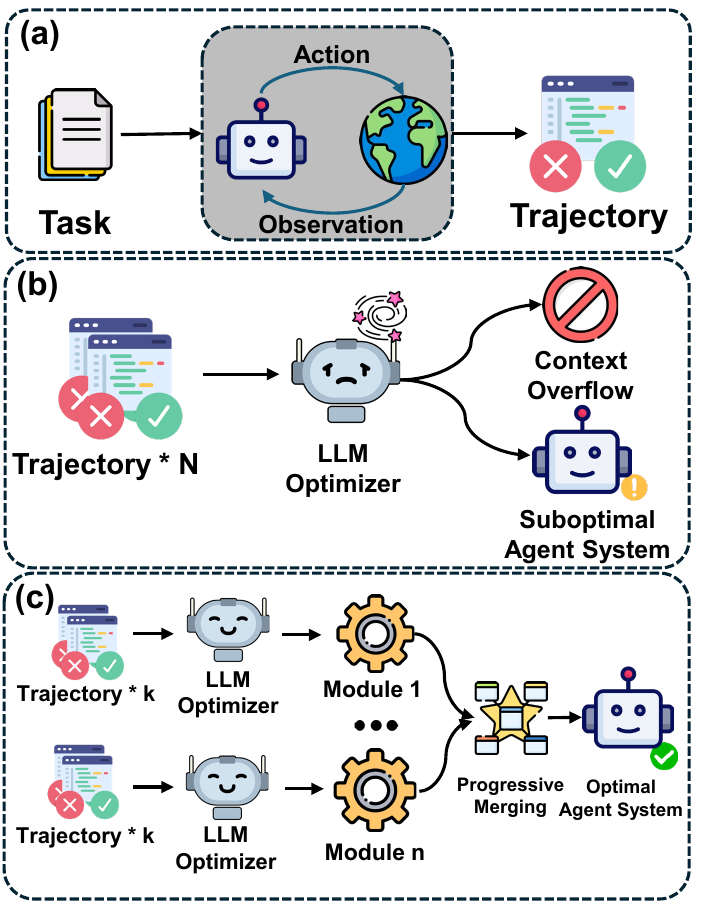}
    \caption{Agent optimization approaches. (a) Basic agent execution process. (b) Traditional all-at-once optimization faces context overflow and inferior performance with large trajectory data. (c) Our method: divide-and-conquer optimization with progressive merging enables scalable processing of large datasets.}
    \label{fig:teaser}
\end{figure}

However, due to the increasing volume of data required for optimizing LLM agentic systems autonomously, directly applying LLM-based optimization approaches encounters a fundamental scalability issue.
Existing methods typically concatenate all execution trajectories on the training data and perform optimization in an all-at-once manner, feeding the entire dataset into the LLM optimizer in a single prompt. 
While this approach works for optimization tasks with small-scale data, it becomes problematic as the data grows.
For instance, in the GAIA benchmark~\citep{mialon2023gaia}, agents normally rely on external tools to collect real-world information and generate lengthy execution traces for subsequent optimization, which is filled with raw documents and complex intermediate reasoning steps, even challenge for human to parse.
This increasing complexity leads to two critical limitations: 
(1) The concatenated trajectories exceed LLM context windows, forcing truncation of valuable optimization signals.
(2) Even when content fits within context windows, LLMs struggle with analyzing long-range dependencies in extensive corpus~\citep{bai2024longbench2,liu2024lost,ni2024xl,ravaut2024context,li2024long}, making it hard for the LLM optimizer to capture subtle patterns and relationships between execution traces. 
As a result, such approaches can produce suboptimal solutions, particularly in complex scenarios where understanding the intricate relationships between different execution trajectories is crucial for improving agent performance.

To address these scalability challenges, we introduce \ours, a framework that enables efficient optimization of LLM-based agentic systems with large-scale data. Specifically,  \ours operates through three components: (1) Task division that breaks down the large training dataset into more manageable subsets, (2) Fine-grained optimization enabling efficient processing of each subset, and (3) Progressive module merging that adaptively combines optimized modules while preserving crucial insights from each subset. This design allows \ours to effectively handle larger optimization tasks while maintaining high-quality results.

We evaluate \ours by optimizing two agent modules: instruction prompts and tools agent could access. Across diverse tasks including ALFWorld~\citep{shridhar2020alfworld}, LogisticsQA, and GAIA~\citep{mialon2023gaia}. 
Agent trained with \ours produces significant performance gains across all datasets, ranging from 8.3\% to 38.1\%, outperforming other optimization methods by 1.6\%-8.6\%. Further analysis reveals that \ours maintains superior performance across varying training dataset sizes, highlighting its scalability and stability. Notably, \ours achieves these improvements while reducing prompt token consumption by 56.3\% and increasing optimization efficiency by 7.6\% compared to conventional all-at-once optimization.

Our contributions are threefold: 
(1) We identify and analyze the scalability limitations in current LLM-based optimization approaches for agentic systems. 
(2) To address the scalability limitation, we propose \ours, a scalable optimization framework that effectively handles large-scale agent optimization through task division and progressive merging. 
(3) We demonstrate \ours's effectiveness across diverse tasks and provide insights into its scalability advantages through comprehensive empirical analysis.

\section{Preliminary}

\subsection{Problem Setup}
\paragraph{LLM Agent Optimizable Modules}
Agentic systems exhibit complex behavioral patterns emerging from multiple factors. A critical insight in designing such systems lies in the decomposition of the agent's parameter space into \textit{modules} that can be independently optimized~\citep{anthropic_effective_agents}. This decomposition enables targeted optimization of specific functional aspects while maintaining global system coherence. Denote the parameter space of agentic system as $\Theta$, which partitions into trainable modules $\{\Theta_i\}_{i=1}^n$ governing distinct behavioral dimensions. Each module must satisfy two key properties to qualify as a modular unit. First, the \textit{trainability} property requires that each module can meaningfully influence the agent's policy gradients when exposed to specific queries. This ensures the module is sufficiently responsive to reward signals during optimization. Second, the \textit{orthogonality} property mandates that parameter gradients across different modules exhibit minimal directional alignment during optimization. Such orthogonality constraint ensures modules encode non-redundant functionalities while guaranteeing each contributes uniquely to performance optimization.

\paragraph{Agentic System Optimization}
An agent interacts with an environment $\mathcal{E}$ by generating a sequence of actions in response to a query. Given parameters $\theta$, the agent's policy determines actions based on the current state of interaction and observation. These actions along with the observations form a trajectory $\tau$ that represents the agent's solution attempt for the given query.

\begin{equation}
\begin{gathered}
a_t \sim \pi(\cdot|a_{1:t},o_{1:t}; \theta), \; o_{t+1} \sim \mathcal{E}(\cdot|a_t),\; \forall t \in [T] \\
\tau = \mathcal{A}(q; \theta) = (o_1, a_1, ..., o_T, a_T)
\end{gathered}
\label{eq:agent}
\end{equation}
The performance is quantified through a loss function $\mathcal{L}$.
Given a distribution $\mathcal{D}$ over query-label pairs $(q, y)$, we aim to find optimal agent parameters that minimize expected loss across the task distribution. The optimization objective is:
\begin{equation}
\label{eq:optim}
\theta^* = \arg\min_{\theta \in \Theta} \mathbb{E}_{(q, y) \sim \mathcal{D}} \left[ \mathcal{L}(\mathcal{A}(q; \theta), y) \right]
\end{equation}

This formulation of optimization via tuning modules provides a unified abstraction for analyzing performance-critical factors in agentic system design. In practice, the modules include prompt for task handling~\citep{wen2024hard,wu2024avatar}, long term memory~\citep{zhang2024survey}, the available toolbox~\citep{zhangoffline}, the weights of backbone LLM~\citep{zeng2024agenttuning,ma2024taco}.

\subsection{Motivation}
\label{motivation}
In LLM-as-optimizer setting, we assume the numeric value of the policy gradient is not accessible in Eq. \ref{eq:optim}. This constraint emerges from a practical reality in modern LLM agent systems - the increasing reliance on proprietary Large Language Models like GPT-4 \citep{openai2023gpt4} and Claude \citep{anthropic2024claude35}, where internal parameters are inaccessible. 

Current approaches that leverage LLM as optimizer typically follow a two-step iterative process: first evaluating modules on training data to collect trajectories and losses, then prompting the LLM optimizer with this information to generate improved modules. While these methods have shown promising results \cite{yang2023large,zhangoffline,cheng2024trace}, they face fundamental scalability challenges that limit their practical applications.

\paragraph{Context window limit.} 
The inherent constraint of LLM context windows is a critical bottleneck in optimization. As training samples grow, the concatenated trajectories and module-loss pairs can exceed the context capacity of even the most capable LLMs. This limitation becomes particularly acute in complex tasks where individual trajectories contain extensive reasoning steps or multi-turn interactions. In such scenarios, even a modest number of samples can overwhelm the context window, severely limiting the LLM optimizer's ability to process comprehensive training data.

\paragraph{Insufficient context utilization.} Even when the content fits within context limits, LLMs can face significant challenges in effectively processing and discovering patterns across extensive collections of trajectories~\citep{ni2024xl,li2024long,bai2024longbench2}. Recent benchmarks on long-form text comprehension and summarization tasks have consistently demonstrated that LLM's performance deteriorates significantly with increasing text length, particularly in processing complex dialogues and lengthy documents~\citep{bai2024longbench2,wu2024longmemeval,ni2024xl,song2024counting,zhang2024bench}. In the context of LLM based optimizers, optimization requires grasping long-range dependencies and analyzing fine-grained details to capture subtle patterns across multiple lengthy samples. This inherent limitation of LLMs can lead to suboptimal module updates that fail to capture the full complexity of the optimization problem, especially in real-world applications where performance depends on understanding both broad patterns and fine-grained details across diverse samples.

\begin{figure*}[htbp]
    \centering
    \includegraphics[width=0.9\linewidth]{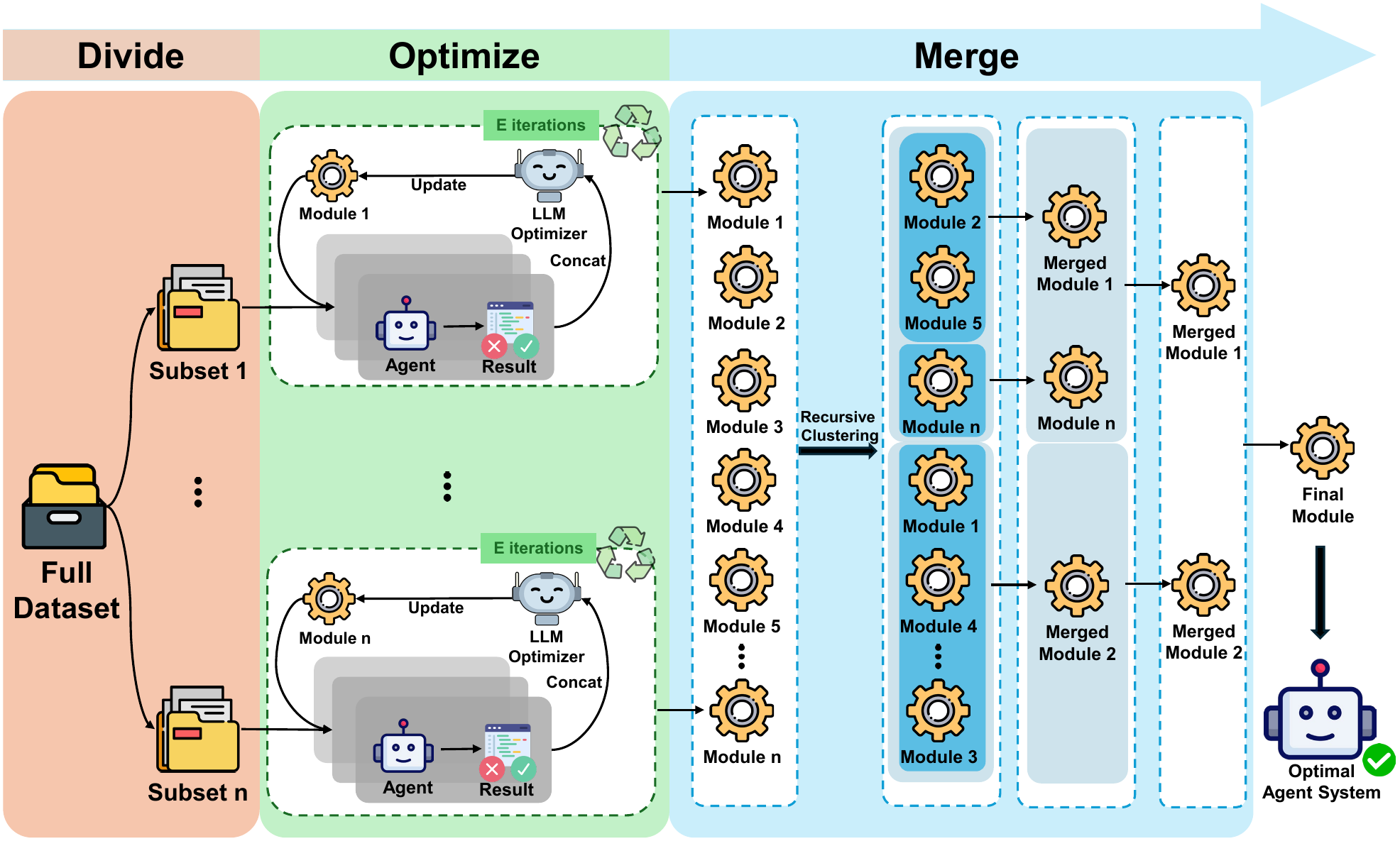}
    \caption{Illustration of \ours's optimization pipeline. The system operates in three stages: (1) Divide: the full dataset is split into manageable subsets, (2) Optimize: parallel optimization is performed on each subset using LLM-based optimization with multiple iterations, and (3) Merge: optimized modules are progressively combined using recursive clustering to produce the final optimal agent system.}
    \label{fig:method}
\end{figure*}

\section{Methods}

\subsection{Overview}
The overall pipeline of \ours is illustrated in Figure \ref{fig:method}. The core concept behind our proposed framework is to divide the large task set into smaller, more manageable subsets and optimize them independently. After we obtain the optimal modules trained on each subsets, we develop an algorithm to progressively merge them into an optimal module.

\subsection{Fine-Grained LLM Agent Optimization}
\label{optim}
\paragraph{Basic Module Optimization}
We begin with describing how we perform agent optimization. The pipeline is illustrated in Algorithm \ref{alg:iter}.
In each epoch, the agent undergoes a three-phase cycle: exploration, evaluation, and optimization. During exploration, the agent interacts with the given task with the current module, generating the solution trajectories. The evaluation phase introduces a post-hoc LLM based evaluator that analyzes these trajectories to determine correctness, identify failures, patterns as well as potential areas for improvement based on the ground truth and trajectory. The evaluations serve as textual gradients to guide the direction for updating the instruction toward better performance. The optimization phase then leverages these insights by feeding the trajectories, textual gradients into an LLM based optimizer, which synthesizes this information to generate an updated module.

\begin{algorithm}[t]
\caption{Module Optimization}
\begin{algorithmic}
\Input Task set $\mathcal{D}$, number of epochs $E$
\Output Optimized module $\theta$
\State $\theta \leftarrow \phi$ \Comment{Start from scratch}
\For{$e \leftarrow 1$ to $E$}
    \State $\mathcal{H} \leftarrow \{\}$ \Comment{Empty trajectory history}
    \For{$(q,y) \in \mathcal{D}$} 
        \State $\tau \leftarrow \mathcal{A}(q; \theta)$ \Comment{Eq. \ref{eq:agent}}
        \State $ r \leftarrow \text{Evaluate}(\tau, y) $ 
        \State $\mathcal{H}.\text{append}((\tau, r))$
    \EndFor
    \State $\theta \leftarrow \text{LLM}_{\textbf{optim}}(\mathcal{H}, \theta)$ \Comment{Update module}
\EndFor
\State \Return $\theta$
\end{algorithmic}
\label{alg:iter}
\end{algorithm}

\paragraph{Divide}
As the number and complexity of task set scales, the length and number of the trajectories can quickly increase, posing challenge to LLM based optimization. To address the issue, we propose a divide-and-conquer based strategy that decomposes the training dataset $D$ into $N$ disjoint subsets $\{D_i\}_{i=1}^N$, and perform optimization on the subsets independently. By operating on smaller, focused subsets, the intuition is to capture subtle patterns and requirements that might be overlooked in global optimization. The process yields $N$ distinct module-loss pairs, each specialized for its respective subset's characteristics.

\paragraph{Progressive Merging}

\begin{algorithm}[t]
\caption{Progressive Module Merging}
\begin{algorithmic}
\Input List $\mathcal{M}=\{(\theta_i, \mathcal{T}_i, p_i)\}$ containing modules, their tasks, and performances
\Output Optimized module $\theta^*$
\Function{ProgressiveMerge}{$\mathcal{M}$, $t$}
    \If{$|\mathcal{M}| \leq t$}
        \State $\theta,p \leftarrow$ Merge($\mathcal{M}$) \Comment{Base: Direct Merge}
        \State \Return $\theta$
    \EndIf
    \State $C \leftarrow$ KMeans($S$,$\sqrt{|\mathcal{M}|}$) \Comment{Adaptive cluster}
    \For{each cluster $c_i \in C$}
        \State $\theta_i, p_i \leftarrow$ ProgressiveMerge($c_i$, $t$)
    \EndFor
    \State \Return Merge($\{\theta_i, p_i\}$)
\EndFunction
\State \Return ProgressiveMerge($\mathcal{M}$, $t$)
\end{algorithmic}

\label{alg:merge}
\end{algorithm}

While decomposition addresses the immediate scalability constraints, it introduces the challenge of integrating $N$ independently optimized modules while preserving their specialized insights. The straightforward approach would be to directly prompt an LLM with all module-performance pairs and generate an updated module. However, such all-at-once merging struggles to effectively process and synthesize patterns across many modules simultaneously, potentially losing the specialized optimizations gained through divided optimization. We propose progressive merging, implemented as a recursive algorithm that controls merging granularity through a cluster size threshold.
Algorithm \ref{alg:merge} shows the process of progressive merging. For a given list of module-performance pairs, we first check if the list size exceeds the threshold. For larger lists, we partition it into $k = \lfloor\sqrt{n}\rfloor$ clusters based on similarities, where $n$ is the number of modules. Each resulting cluster then undergoes recursive merging. When a cluster's size falls below the threshold, we merge its modules by prompting an LLM with the module contents and their corresponding performance statistics. After each merge operation, we evaluate the merged module's performance through validating on the combined task set from all constituent modules. The recursive process naturally builds a bottom-up merging tree, where each internal node represents a validated merge of its children's modules. This controlled, progressive approach ensures that each merge operation stays within LLM context limits while capturing intricate relationships between similar modules, ultimately enabling efficient optimization of large-scale agentic systems.

\section{Evaluations}
\subsection{Experiment Setup}
We evaluate \ours by optimizing two different modules of the agentic system: instructions and tools. With proper instructions on the guidelines for the tasks, the agent can comprehend the scenario and solve it with ease~\citep{fu2024autoguide,chen2024automanual,wu2024avatar,zhao2024expel}. The tools expand the action space available to the agent, functioning as specialized modules that enable specific capabilities. Optimizing the tool configuration directly impacts the agent's ability to execute complex tasks efficiently and accurately.

\paragraph{Datasets} We conduct experiments on three different benchmarks to study the performance of \ours.
\begin{itemize}[leftmargin=*,
                itemsep=0em,     
                parsep=0em,    
                topsep=0em,    
                partopsep=0em]   
    \item \textbf{ALFWorld}~\citep{shridhar2020alfworld} is a text-based benchmark in which the agent is tasked with performing household tasks. Given a high-level objective, the agent needs to interact with the virtual environment and perform actions through natural language to finish the task. We randomly select 60 tasks from the training datasets (10 for each task type), and use the unseen set containing 134 tasks as test set. The benchmark contains 6 types of tasks, we set the number of agent optimizers to 6, with each agent optimizer optimizing each type of task. We report the success rate on different types of tasks and the overall success rate. 
    \item \textbf{LogisticsQA} is our own curated benchmark. The dataset consists of UBL format shipping invoice documents from real world scenarios. The agent is tasked to understand and extract the transport reference number from the document. The dataset contains 267 document instances. For further details of the dataset, please refer to Appendix ~\ref{anon}. We randomly select 48 documents for training, the remaining 219 for testing. We set the number of agent optimizers to 8, each performs optimization on randomly split 6 tasks. A task is considered successful if the agent's answer is an exact match with the ground truth.
    \item \textbf{GAIA}~\citep{mialon2023gaia} is a benchmark designed to test the capability of agents as general assistants. It encompasses tasks from different domains such as file browsing, web searching and scraping, making it a perfect testbed for benchmarking agent's tool usage capability as well as the quality of the toolbox. We utilize 36 tasks from the training set and evaluate on 60 tasks. The optimization is distributed across 4 optimizers, each handling a distinct subset of tasks.

\end{itemize}

\paragraph{Baselines for Comparison.} We compare performance with different agent optimization methods: (1) \textbf{All-at-once} optimization represents the conventional approach of performing agent optimization on the whole training set using the algorithm illustrated in Section \ref{optim}; (2) \textbf{Batch-wise} optimization employs a fixed-size batching strategy, splitting the training dataset into predetermined chunks and performing optimization sequentially on each task batch within an epoch; (3) \textbf{Bootstrapping} optimization implements a stochastic approach, sampling task batches from the training dataset with replacement.

\paragraph{Implementation details.} We optimize the instructions for the agent on ALFWorld and LogisticsQA, and optimize the tools on GAIA. For ALFWorld, we leverage gpt-4o-mini as the backbone for the agent and evaluator, and gpt-4o for optimization and merging. For LogisticsQA and GAIA, we use gpt-4o in the whole process. For a fair comparison, all methods use the same number of optimization steps.

\subsection{Main Results}
\label{main}

\begin{table*}[t]
\centering
\resizebox{\textwidth}{!}{%
\begin{tabular}{c|ccccccc|c|c}
\hhline{=:=======:=:=}
\rowcolor{light-gray0} 
\cellcolor{light-gray0}  &
  \multicolumn{7}{c|}{\cellcolor{light-gray0} ALFWorld} &
  \cellcolor{light-gray0}  &
  \cellcolor{light-gray0}  \\
\rowcolor{light-gray0}  
\multirow{-2}{*}{\cellcolor{light-gray0} Methods} &
  Pick &
  Clean &
  Heat &
  Cool &
  Look &
  Pick Two &
  Average &
  \multirow{-2}{*}{\cellcolor{light-gray0} LogisticQA} &
  \multirow{-2}{*}{\cellcolor{light-gray0} GAIA} \\ 
\hline
Vanilla Agent            & 69.4 & 50.5 & 65.2 & 20.6 & 31.5 & 21.6 & 45.5 & 36.3 & 15.0 \\ 
\hline
All-at-once          & 90.2  &  72.6  & 78.3  &  78.6 &  66.7  &  55.9 &  75.0 & 52.1* & 21.7* \\
Batch-wise       &  77.1    &  71.0    &  67.4    &  64.3    &  \textbf{86.1}    &  \textbf{73.5}    &  72.8    & 55.7 & 10.0 \\
Bootstrapping       & \textbf{91.7}     &  77.4    &  73.9    &  74.6  & 87.0     &  41.2    &  75.6    & 62.6  & 20.0 \\
\rowcolor{light-gray} 
FGO &  90.2   &  \textbf{83.8}   & \textbf{87.0}  &  \textbf{88.9}   &   \textbf{86.1}    &  62.7  &  \textbf{83.6}   & \textbf{64.8 }& \textbf{23.3} \\ 
\hhline{=:=======:=:=}
\end{tabular}%
}
\caption{Performance of the optimized agent using different optimization methods on ALFWorld, LogisticQA and GAIA. The best results are in bold. * denotes that we encounter context window exceeded error during optimization and have to trim the number of trajectory reward pairs sent to the LLM optimizer.}
\label{tab:main}
\end{table*}

\paragraph{Finding 1: FGO demonstrates superior optimization performance across multiple domains.}
We present the optimized agent's performance on different benchmarks in Table \ref{tab:main}. For the majority of the tasks, the agent demonstrates performance gain in most cases after optimization. This highlights how targeted prompt and tool refinement can significantly enhance LLM agent capabilities. Among the optimization methods, \ours achieves the most performance boost in all cases, with gains ranging from 8.3\% to 38.1\% compared to the vanilla agents.

\paragraph{Finding 2: Progressive merging effectively preserves task-specific patterns while achieving global optimal.}
The superior performance of \ours can be attributed to its divide-and-conquer-based methodology. The All-at-once approach processes the entire training dataset simultaneously, requiring the LLM optimizer to learn from trajectories across the complete dataset. This leads to suboptimal performance due to the optimizer's difficulty in processing complex patterns in long corpus, as evidenced by the suboptimal performance on ALFWorld subtasks. Alternative methods like bootstrapping optimization and batch-wise optimization demonstrate strong performance in specific categories, but fail to maintain consistent performance across the task spectrum. Their batch-wise optimization approach introduces instability in the training process, as the LLM optimizer encounters different data distributions in successive iterations, potentially compromising previously learned patterns. In contrast, \ours overcomes these limitations through its systematic merging of independently optimized instructions and tools. By first optimizing subset-specific instructions and tools and then progressively merging them, \ours can preserve task-specific patterns while building toward global optimization. We further examine the implication of merging on \ours performance in Section \ref{ablation}.

\subsection{Further Analysis}

\paragraph{Finding 3: \ours demonstrates extraordinary scalability.}
\begin{figure}[t]
    \centering
    \includegraphics[width=\linewidth]{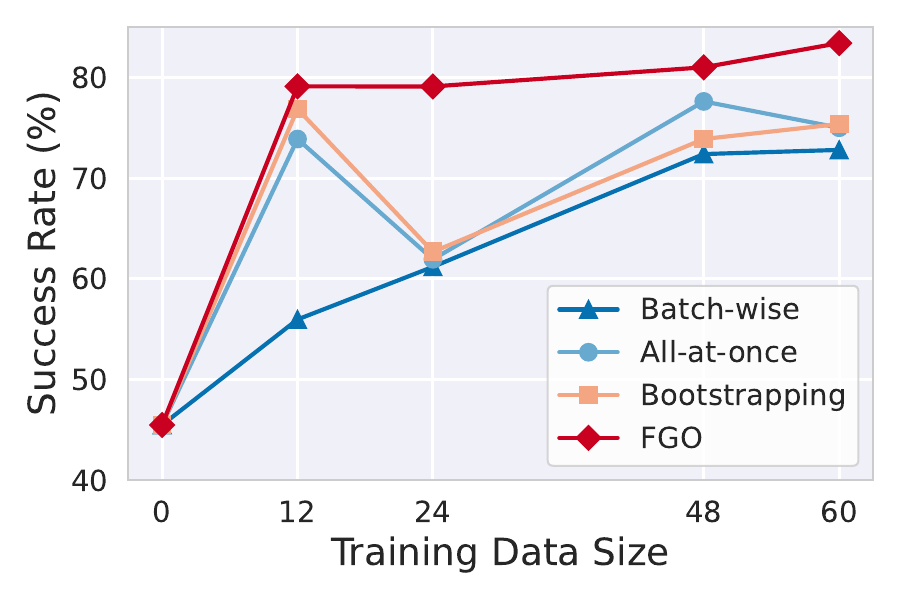}
    \caption{Analysis of the number of training tasks. We run optimization on varied training dataset sizes and plot the performance. \ours achieves best performance in all training settings, and demonstrate a steady increase as the training taskset size increases.}
    \label{fig:ablation-num-samples}
\end{figure}

We evaluated how training data volume affects optimization performance on ALFWorld. As shown in Figure \ref{fig:ablation-num-samples}, \ours maintains stable performance across all dataset sizes, with consistent improvements as training samples increase. While batch-wise optimization shows similar training accuracy in low-data settings, it yields lower performance compared to bootstrapping optimization, indicating poorer generalization. This aligns with established machine learning principles where bootstrapping enhances generalization~\citep{breiman1996bagging}. Additionally, All-at-once optimization proves impractical for LogisticsQA and GAIA due to their extensive document lengths (>3,000 tokens) and complex solution trajectories exceeding LLM context windows, validating the need for our scalable approach.

\paragraph{Finding 4: \ours achieves an optimal balance between token cost, efficiency and performance.}

\begin{figure*}[htbp]
	\centering
	\begin{subfigure}{0.32\linewidth}
		\centering
		\includegraphics[width=\linewidth]{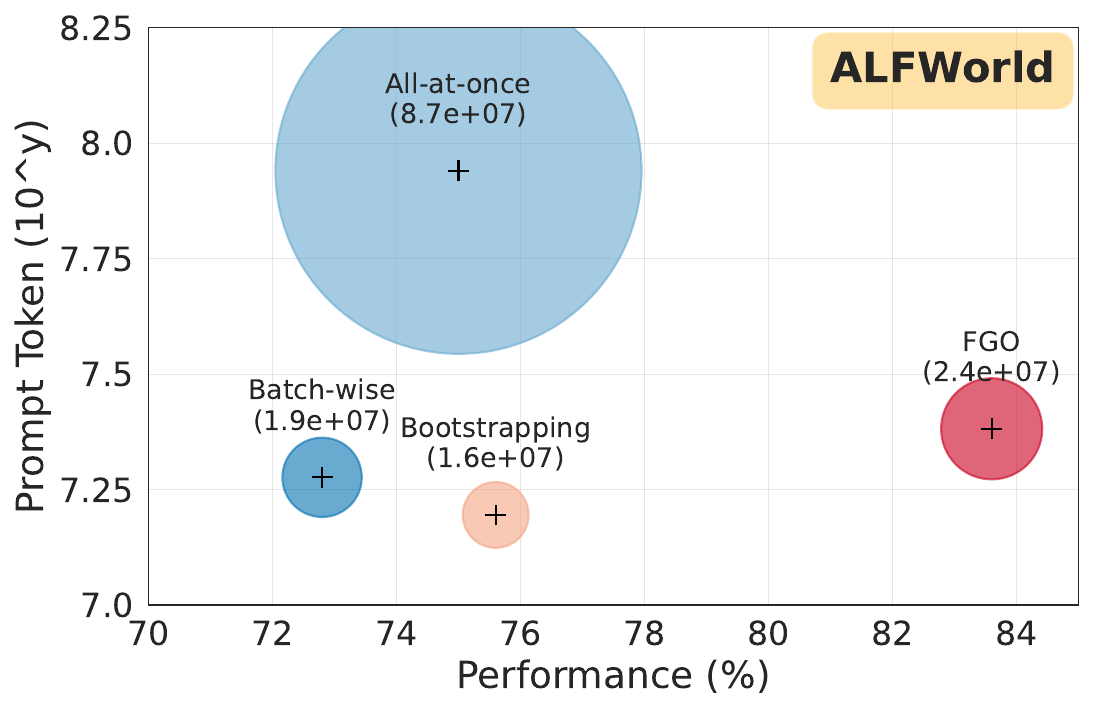}
	\end{subfigure}
	\centering
	\begin{subfigure}{0.32\linewidth}
		\centering
		\includegraphics[width=\linewidth]{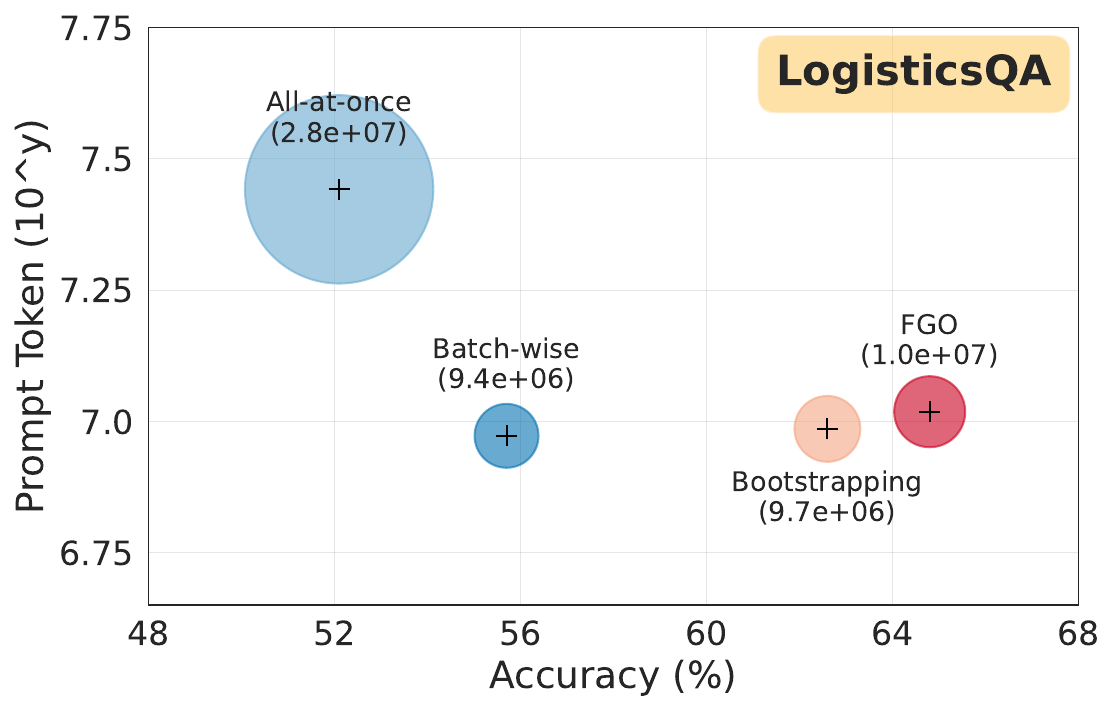}
	\end{subfigure}
	\centering
	\begin{subfigure}{0.32\linewidth}
		\centering
		\includegraphics[width=\linewidth]{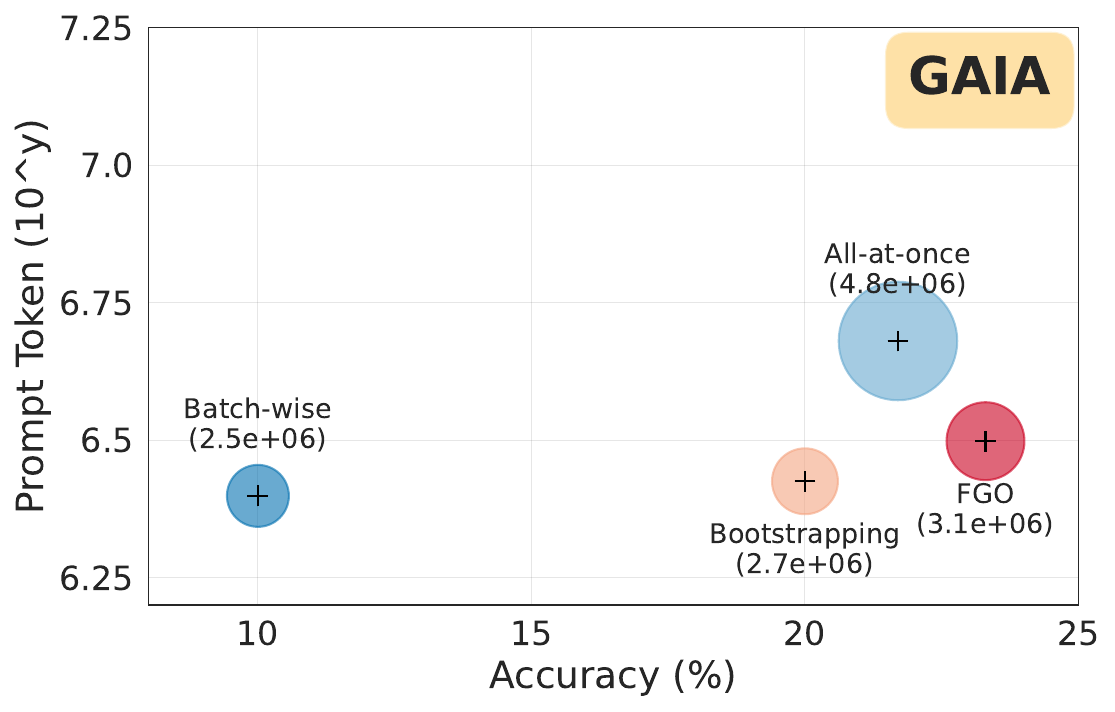}
	\end{subfigure}
	\caption{Comparison of prompt token efficiency across different optimization methods on ALFWorld, LogisticsQA, and GAIA. Each panel plots the trained agent's performance against the total prompt tokens consumed during optimization. Circle diameters are proportional to the optimization token consumption, with crosses (+) indicating circle centers.}
    \label{fig:cost}
\end{figure*}
We visualize the relationship between prompt token used for optimization and the performance after training with different optimization methods in Figure \ref{fig:cost}, and outline the time to train and performance in Figure \ref{fig:cost_time}.
In terms of token cost, \ours requires larger number of prompt tokens compared to Batch-wise optimization and Boostrapping optimization. This is because the merging process requires evaluating on the combined task set from all the constituent modules. This is a sacrifice in exchange for accurately modeling the merged module's capability in order to generate more accurate modules in the merging process. In contrast, All-at-once prompts the LLM optimizer with the whole list of trajectories and losses, leading to the largest token consumption requirements than other methods.
In terms of efficiency, \ours can perform optimization in parallel and gather the optimized modules at once, which is an unique advantage compared to the sequential training methods.


\subsection{Ablation Study}
\label{ablation}
We investigate the following questions to understand the impact of different components and hyperparameters in \ours:

\paragraph{How does progressive merging and choice of clustering algorithm affect performance?}

To analyze the impact of progressive merging and evaluate different clustering algorithms in the merging process, we conduct experiments on the ALFWorld benchmark. We first establish a baseline by removing the progressive merging entirely, and instead prompting an LLM to generate the final module directly from the module-performance pairs obtained from divided optimization. We then evaluate the effect of different clustering algorithms by fixing the independently optimized modules and changing the clustering method to Spectral clustering and Bisect K-Means.
We report the average and best of three runs in Table \ref{ablation: cluster}. Without progressive merging, the method achieves a 73.1\% average success rate, demonstrating that even basic merging provides substantial improvement over the vanilla agent. In comparison, the introduction of progressive merging significantly boosts performance. Regardless of the clustering algorithm employed during merging, the final performances all demonstrate consistent improvement to no merging. This consistency suggests that the progressive nature of the merging strategy, rather than the specific clustering algorithm, is the key driver of improvement.

\begin{table}[]
\centering
\resizebox{\columnwidth}{!}{%
\begin{tabular}{c|c|cc}
\hline
\rowcolor{light-gray0} 
\cellcolor{light-gray0}  & \cellcolor{light-gray0}         & \multicolumn{2}{c}{\cellcolor{light-gray0} ALFWorld} \\
\rowcolor{light-gray0}  
\multirow{-2}{*}{\cellcolor{light-gray0} Methods} &
  \multirow{-2}{*}{\cellcolor{light-gray0} \begin{tabular}[c]{@{}c@{}}Cluster\\ Algorithm\end{tabular}} &
  Avg of 3 &
  Best of 3 \\ \hline
Vanilla                     & -                               &      45.5              &      61.9              \\ \hline
                         & None                            &    73.1                &       84.3                 \\
                         & Spectral             &      81.6            &      89.6              \\
                         & Bisect K-Means                  &    80.1                &        \textbf{91.0}              \\
\multirow{-4}{*}{\ours}   & K-Means & \textbf{83.6} & 89.6  \\ \hline
\end{tabular}%
}
\caption{Ablation study on the effects of clustering algorithms used. "None" means we skip the clustering step and directly merge the optimized modules.}
\label{ablation: cluster}
\end{table}

\paragraph{Does the division of training data affect performance?}
To examine the robustness of \ours, we compare our default category-based partitioning against random partitioning of training tasks in ALFWorld. As shown in Table \ref{tab:abla-split}, while random partitioning shows a slight performance drop, the system still maintains strong performance thanks to the progressive merging process, which effectively combines optimization insights across partitions. This demonstrates that \ours's performance remains robust even with suboptimal partitioning strategies.

\begin{table}[t]
\centering
\resizebox{0.7\columnwidth}{!}{%
\begin{tabular}{c|cc}
\hline
\rowcolor{light-gray0} 
\cellcolor{light-gray0}                         & \multicolumn{2}{c}{\cellcolor{light-gray0} ALFWorld} \\
\rowcolor{light-gray0}  
\multirow{-2}{*}{\cellcolor{light-gray0} Partition} & Avg of 3                 & Best of 3                 \\ \hline
Random                                          &     80.3                  &  88.1                     \\
Category                                       &       \textbf{83.6}            &    \textbf{89.6 }                 \\ \hline
\end{tabular}%
}
\caption{Ablation on the data partitioning strategy. 'Category' denotes we partition the training data according to the task type. }
\label{tab:abla-split}
\end{table}

\paragraph{How does the number of divided subsets affect performance?}

\begin{figure}[t]
    \centering
    \includegraphics[width=0.8\columnwidth]{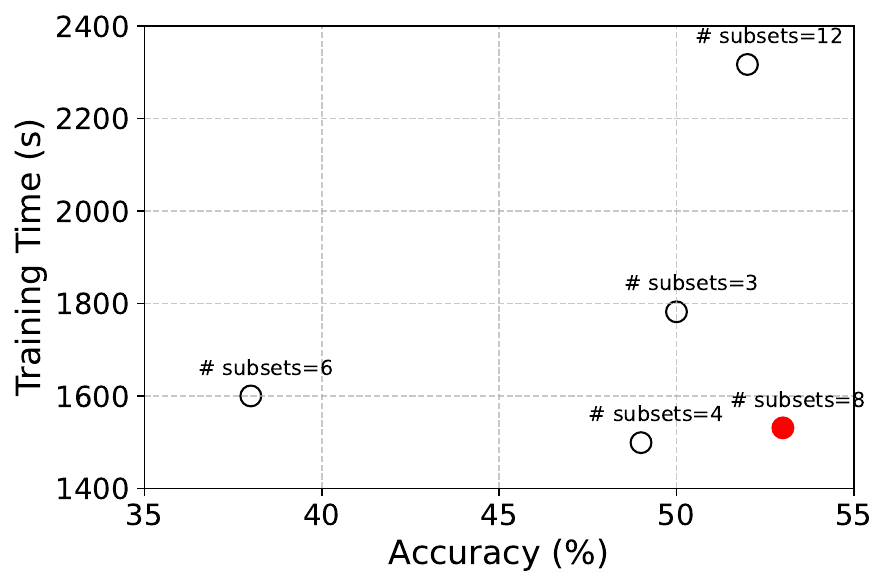}
    \caption{Ablation study on the number of divided subsets. Most parameter settings achieve similar performance, with varying time for optimization.}
    \label{fig:num-agents}
\end{figure}

To answer this question, we conduct ablation study on the number of independent agent optimizers. We trained agents on LogisticsQA and set the number of divided subsets to 3, 4, 6, 8, 12. Due to the high cost in running \texttt{gpt-4o}, we sampled 100 tasks from the test set and validate the optimized agent's performance respectively. We plot the relationship between performance and training time in Figure \ref{fig:num-agents}. The results suggest that the choice of agent number primarily impacts computational efficiency rather than optimization quality.

\paragraph{How does performance change with respect to backbone LLM?}
Finally, to ensure \ours well generalizes to different backbones, we change the optimizer backbone to o3-mini and observe the metrics. We report the token consumption for training, wall-clock time, and performance in Table \ref{tab:abla-llm}. In line with the main results, \ours maintains strong performance while reaching most efficiency, with slightly overhead in token consumption.

\begin{table}[]
\centering
\resizebox{\columnwidth}{!}{%
\begin{tabular}{c c c c c}
\hline
\rowcolor{light-gray0}
Method & \# Tokens ($10^7$) & Time (s) & Avg of 3 & Best of 3 \\
\hline
All-at-once   & 8.15 & 7583 & 87.8 & 93.3 \\
Batch-wise    & 1.59 & 2969 & 83.1 & 92.3 \\
Bootstrapping & \textbf{1.34} & 2521 & 88.8 & 95.0 \\
\rowcolor{light-gray}
FGO           & 1.97 & \textbf{2142} & \textbf{89.3} & \textbf{95.5} \\
\hline
\end{tabular}%
}
\caption{Ablation on the optimizer backbone. We leverage o3-mini as the backbone for optimization, and report the metrics. Best result is in bold.}
\label{tab:abla-llm}
\end{table}

\section{Related Work}
\label{others}

\paragraph{LLM as Optimizer.}
LLMs are increasingly used as a blackbox optimizer for different LLM systems. In prompt optimization, LLM is leveraged to automously maximizing LLM's performance to novel tasks without expensive model tuning~\citep{zhou2022large,pryzant2023automatic,cheng2023black,prasad2022grips,opsahl2024optimizing,khattab2024dspy}. In the realm of in-context learning~\citep{min2021metaicl,dong2022survey,brown2020language}, by automatically retrieving demonstrations from training set~\citep{zhao21calibrate,lu2021fantastically,liu2021makes} or from adaptively annotated samples by LLM~\citep{zhang2023ideal,wu2022self,su2022selective}, prompt with autonomously selected in-context examples can reach performance better can human crafted prompts. LLM based optimizers are also used as a meta-optimizers to debug and improve an LLM based system~\citep{zelikman2023self,yin2024g,zhang2025agent,song2024adaptive}.

\paragraph{Automated Agentic System Design.} There has been efforts in exploring inference time performance boost since the emergence of Large Language Models~\citep{shinn2024reflexion, madaan2024self,yao2023react,yao2024tree,wei2022chain,guo2024embodied,wu2024autogen}. Recent works have extended this paradigm to agentic systems. 
Some works represent and learn the optimal workflow of agentic systems in the form of complex graphs~\citep{zhuge2024language,wu2024stateflow}, code~\citep{hu2024automated}, and trees~\citep{zhang2024aflow} to improve the system's performance on complex tasks, while others learns reusable tools~\citep{zhangoffline,cai2023large,qian2023creator,yuan2023craft} and experience~\citep{zhao2024expel,wang2024agent} for agentic systems.

\section{Conclusion}
In this paper, we addressed the scalability challenges in LLM-based agent optimization by introducing \ours, a framework that effectively processes large-scale execution trajectories through task division, fine-grained optimization, and progressive module merging. Our evaluation across multiple dataset demonstrates consistent performance improvements. \ours reaches an optimal balance between performance, efficiency and token consumption.

\section*{Limitations}
The merging process introduces computational overhead, as it requires to back test the merged module on the merged training dataset, resulting in larger token cost compared to Batch-wise optimization and Bootstrappingoptimization. In future works, we attempt to leverage LLM to predict the performance of the merged module using in-context learning, or approximate the performance using Bayesian methods.

\bibliography{main}

\begin{thebibliography}{70}
\providecommand{\natexlab}[1]{#1}

\bibitem[{Abuelsaad et~al.(2024)Abuelsaad, Akkil, Dey, Jagmohan, Vempaty, and Kokku}]{abuelsaad2024agent}
Tamer Abuelsaad, Deepak Akkil, Prasenjit Dey, Ashish Jagmohan, Aditya Vempaty, and Ravi Kokku. 2024.
\newblock Agent-e: From autonomous web navigation to foundational design principles in agentic systems.
\newblock \emph{arXiv preprint arXiv:2407.13032}.

\bibitem[{Anthropic(2024{\natexlab{a}})}]{anthropic_effective_agents}
Anthropic. 2024{\natexlab{a}}.
\newblock Building effective agents.
\newblock \url{https://www.anthropic.com/research/building-effective-agents}.

\bibitem[{Anthropic(2024{\natexlab{b}})}]{anthropic2024claude35}
Anthropic. 2024{\natexlab{b}}.
\newblock \href {https://assets.anthropic.com/m/1cd9d098ac3e6467/original/Claude-3-Model-Card-October-Addendum.pdf} {Model card addendum: Claude 3.5 haiku and upgraded claude 3.5 sonnet}.

\bibitem[{Bai et~al.(2024)Bai, Tu, Zhang, Peng, Wang, Lv, Cao, Xu, Hou, Dong, Tang, and Li}]{bai2024longbench2}
Yushi Bai, Shangqing Tu, Jiajie Zhang, Hao Peng, Xiaozhi Wang, Xin Lv, Shulin Cao, Jiazheng Xu, Lei Hou, Yuxiao Dong, Jie Tang, and Juanzi Li. 2024.
\newblock Longbench v2: Towards deeper understanding and reasoning on realistic long-context multitasks.
\newblock \emph{arXiv preprint arXiv:2412.15204}.

\bibitem[{Breiman(1996)}]{breiman1996bagging}
Leo Breiman. 1996.
\newblock Bagging predictors.
\newblock \emph{Machine learning}, 24:123--140.

\bibitem[{Brown(2020)}]{brown2020language}
Tom~B Brown. 2020.
\newblock Language models are few-shot learners.
\newblock \emph{arXiv preprint arXiv:2005.14165}.

\bibitem[{Cai et~al.(2023)Cai, Wang, Ma, Chen, and Zhou}]{cai2023large}
Tianle Cai, Xuezhi Wang, Tengyu Ma, Xinyun Chen, and Denny Zhou. 2023.
\newblock Large language models as tool makers.
\newblock \emph{arXiv preprint arXiv:2305.17126}.

\bibitem[{Chen et~al.(2024)Chen, Li, Yang, Yu, Lin, and He}]{chen2024automanual}
Minghao Chen, Yihang Li, Yanting Yang, Shiyu Yu, Binbin Lin, and Xiaofei He. 2024.
\newblock Automanual: Generating instruction manuals by llm agents via interactive environmental learning.
\newblock \emph{arXiv preprint arXiv:2405.16247}.

\bibitem[{Cheng et~al.(2024)Cheng, Nie, and Swaminathan}]{cheng2024trace}
Ching-An Cheng, Allen Nie, and Adith Swaminathan. 2024.
\newblock Trace is the next autodiff: Generative optimization with rich feedback, execution traces, and llms.
\newblock \emph{arXiv preprint arXiv:2406.16218}.

\bibitem[{Cheng et~al.(2023)Cheng, Liu, Zheng, Ke, Wang, Dong, Tang, and Huang}]{cheng2023black}
Jiale Cheng, Xiao Liu, Kehan Zheng, Pei Ke, Hongning Wang, Yuxiao Dong, Jie Tang, and Minlie Huang. 2023.
\newblock Black-box prompt optimization: Aligning large language models without model training.
\newblock \emph{arXiv preprint arXiv:2311.04155}.

\bibitem[{Dong et~al.(2022)Dong, Li, Dai, Zheng, Wu, Chang, Sun, Xu, and Sui}]{dong2022survey}
Qingxiu Dong, Lei Li, Damai Dai, Ce~Zheng, Zhiyong Wu, Baobao Chang, Xu~Sun, Jingjing Xu, and Zhifang Sui. 2022.
\newblock A survey on in-context learning.
\newblock \emph{arXiv preprint arXiv:2301.00234}.

\bibitem[{Fu et~al.(2024)Fu, Kim, Kim, Sohn, Logeswaran, Bae, and Lee}]{fu2024autoguide}
Yao Fu, Dong-Ki Kim, Jaekyeom Kim, Sungryull Sohn, Lajanugen Logeswaran, Kyunghoon Bae, and Honglak Lee. 2024.
\newblock Autoguide: Automated generation and selection of context-aware guidelines for large language model agents.
\newblock In \emph{The Thirty-eighth Annual Conference on Neural Information Processing Systems}.

\bibitem[{Guo et~al.(2024)Guo, Huang, Liu, Fan, V{\'e}lez, Wu, Wang, Griffiths, and Wang}]{guo2024embodied}
Xudong Guo, Kaixuan Huang, Jiale Liu, Wenhui Fan, Natalia V{\'e}lez, Qingyun Wu, Huazheng Wang, Thomas~L Griffiths, and Mengdi Wang. 2024.
\newblock Embodied llm agents learn to cooperate in organized teams.
\newblock \emph{arXiv preprint arXiv:2403.12482}.

\bibitem[{Hu et~al.(2024{\natexlab{a}})Hu, Lu, and Clune}]{hu2024automated}
Shengran Hu, Cong Lu, and Jeff Clune. 2024{\natexlab{a}}.
\newblock Automated design of agentic systems.
\newblock \emph{arXiv preprint arXiv:2408.08435}.

\bibitem[{Hu et~al.(2024{\natexlab{b}})Hu, Zhao, Wei, Chai, Ma, Wang, Wang, Su, Xu, Zhu et~al.}]{hu2024infiagent}
Xueyu Hu, Ziyu Zhao, Shuang Wei, Ziwei Chai, Qianli Ma, Guoyin Wang, Xuwu Wang, Jing Su, Jingjing Xu, Ming Zhu, et~al. 2024{\natexlab{b}}.
\newblock Infiagent-dabench: Evaluating agents on data analysis tasks.
\newblock \emph{arXiv preprint arXiv:2401.05507}.

\bibitem[{Jimenez et~al.(2023)Jimenez, Yang, Wettig, Yao, Pei, Press, and Narasimhan}]{jimenez2023swe}
Carlos~E Jimenez, John Yang, Alexander Wettig, Shunyu Yao, Kexin Pei, Ofir Press, and Karthik Narasimhan. 2023.
\newblock Swe-bench: Can language models resolve real-world github issues?
\newblock \emph{arXiv preprint arXiv:2310.06770}.

\bibitem[{Jing et~al.(2024)Jing, Huang, Wang, Yao, Yu, Ma, Zhang, Du, and Yu}]{jing2024dsbench}
Liqiang Jing, Zhehui Huang, Xiaoyang Wang, Wenlin Yao, Wenhao Yu, Kaixin Ma, Hongming Zhang, Xinya Du, and Dong Yu. 2024.
\newblock Dsbench: How far are data science agents to becoming data science experts?
\newblock \emph{arXiv preprint arXiv:2409.07703}.

\bibitem[{Khattab et~al.(2024)Khattab, Singhvi, Maheshwari, Zhang, Santhanam, A, Haq, Sharma, Joshi, Moazam, Miller, Zaharia, and Potts}]{khattab2024dspy}
Omar Khattab, Arnav Singhvi, Paridhi Maheshwari, Zhiyuan Zhang, Keshav Santhanam, Sri~Vardhamanan A, Saiful Haq, Ashutosh Sharma, Thomas~T. Joshi, Hanna Moazam, Heather Miller, Matei Zaharia, and Christopher Potts. 2024.
\newblock \href {https://openreview.net/forum?id=sY5N0zY5Od} {{DSP}y: Compiling declarative language model calls into state-of-the-art pipelines}.
\newblock In \emph{The Twelfth International Conference on Learning Representations}.

\bibitem[{Khattab et~al.(2023)Khattab, Singhvi, Maheshwari, Zhang, Santhanam, Vardhamanan, Haq, Sharma, Joshi, Moazam et~al.}]{khattab2023dspy}
Omar Khattab, Arnav Singhvi, Paridhi Maheshwari, Zhiyuan Zhang, Keshav Santhanam, Sri Vardhamanan, Saiful Haq, Ashutosh Sharma, Thomas~T Joshi, Hanna Moazam, et~al. 2023.
\newblock Dspy: Compiling declarative language model calls into self-improving pipelines.
\newblock \emph{arXiv preprint arXiv:2310.03714}.

\bibitem[{Li et~al.(2024)Li, Zhang, Do, Yue, and Chen}]{li2024long}
Tianle Li, Ge~Zhang, Quy~Duc Do, Xiang Yue, and Wenhu Chen. 2024.
\newblock Long-context llms struggle with long in-context learning.
\newblock \emph{arXiv preprint arXiv:2404.02060}.

\bibitem[{Liu et~al.(2021)Liu, Shen, Zhang, Dolan, Carin, and Chen}]{liu2021makes}
Jiachang Liu, Dinghan Shen, Yizhe Zhang, Bill Dolan, Lawrence Carin, and Weizhu Chen. 2021.
\newblock What makes good in-context examples for gpt-$3 $?
\newblock \emph{arXiv preprint arXiv:2101.06804}.

\bibitem[{Liu et~al.(2024)Liu, Lin, Hewitt, Paranjape, Bevilacqua, Petroni, and Liang}]{liu2024lost}
Nelson~F Liu, Kevin Lin, John Hewitt, Ashwin Paranjape, Michele Bevilacqua, Fabio Petroni, and Percy Liang. 2024.
\newblock Lost in the middle: How language models use long contexts.
\newblock \emph{Transactions of the Association for Computational Linguistics}, 12:157--173.

\bibitem[{Lu et~al.(2021)Lu, Bartolo, Moore, Riedel, and Stenetorp}]{lu2021fantastically}
Yao Lu, Max Bartolo, Alastair Moore, Sebastian Riedel, and Pontus Stenetorp. 2021.
\newblock Fantastically ordered prompts and where to find them: Overcoming few-shot prompt order sensitivity.
\newblock \emph{arXiv preprint arXiv:2104.08786}.

\bibitem[{Ma et~al.(2024)Ma, Zhang, Liu, Zhang, Tan, Shu, Niebles, Heinecke, Wang, Xiong et~al.}]{ma2024taco}
Zixian Ma, Jianguo Zhang, Zhiwei Liu, Jieyu Zhang, Juntao Tan, Manli Shu, Juan~Carlos Niebles, Shelby Heinecke, Huan Wang, Caiming Xiong, et~al. 2024.
\newblock Taco: Learning multi-modal action models with synthetic chains-of-thought-and-action.
\newblock \emph{arXiv preprint arXiv:2412.05479}.

\bibitem[{Madaan et~al.(2024)Madaan, Tandon, Gupta, Hallinan, Gao, Wiegreffe, Alon, Dziri, Prabhumoye, Yang et~al.}]{madaan2024self}
Aman Madaan, Niket Tandon, Prakhar Gupta, Skyler Hallinan, Luyu Gao, Sarah Wiegreffe, Uri Alon, Nouha Dziri, Shrimai Prabhumoye, Yiming Yang, et~al. 2024.
\newblock Self-refine: Iterative refinement with self-feedback.
\newblock \emph{Advances in Neural Information Processing Systems}, 36.

\bibitem[{Mialon et~al.(2023)Mialon, Fourrier, Swift, Wolf, LeCun, and Scialom}]{mialon2023gaia}
Gr{\'e}goire Mialon, Cl{\'e}mentine Fourrier, Craig Swift, Thomas Wolf, Yann LeCun, and Thomas Scialom. 2023.
\newblock Gaia: a benchmark for general ai assistants.
\newblock \emph{arXiv preprint arXiv:2311.12983}.

\bibitem[{Min et~al.(2021)Min, Lewis, Zettlemoyer, and Hajishirzi}]{min2021metaicl}
Sewon Min, Mike Lewis, Luke Zettlemoyer, and Hannaneh Hajishirzi. 2021.
\newblock Metaicl: Learning to learn in context.
\newblock \emph{arXiv preprint arXiv:2110.15943}.

\bibitem[{Ni et~al.(2024)Ni, Cai, Wei, Wang, Yin, and Li}]{ni2024xl}
Xuanfan Ni, Hengyi Cai, Xiaochi Wei, Shuaiqiang Wang, Dawei Yin, and Piji Li. 2024.
\newblock X$\text{L}^{2}$ bench: A benchmark for extremely long context understanding with long-range dependencies.
\newblock \emph{arXiv preprint arXiv:2404.05446}.

\bibitem[{OpenAI(2023)}]{openai2023gpt4}
OpenAI. 2023.
\newblock \href {https://cdn.openai.com/papers/gpt-4-system-card.pdf} {Gpt-4 system card}.

\bibitem[{Opsahl-Ong et~al.(2024)Opsahl-Ong, Ryan, Purtell, Broman, Potts, Zaharia, and Khattab}]{opsahl2024optimizing}
Krista Opsahl-Ong, Michael~J Ryan, Josh Purtell, David Broman, Christopher Potts, Matei Zaharia, and Omar Khattab. 2024.
\newblock \href {https://doi.org/10.18653/v1/2024.emnlp-main.525} {Optimizing instructions and demonstrations for multi-stage language model programs}.
\newblock In \emph{Proceedings of the 2024 Conference on Empirical Methods in Natural Language Processing}, pages 9340--9366, Miami, Florida, USA. Association for Computational Linguistics.

\bibitem[{Pan et~al.(2024)Pan, Wang, Neubig, Jaitly, Ji, Suhr, and Zhang}]{pan2024training}
Jiayi Pan, Xingyao Wang, Graham Neubig, Navdeep Jaitly, Heng Ji, Alane Suhr, and Yizhe Zhang. 2024.
\newblock Training software engineering agents and verifiers with swe-gym.
\newblock \emph{arXiv preprint arXiv:2412.21139}.

\bibitem[{Prasad et~al.(2022)Prasad, Hase, Zhou, and Bansal}]{prasad2022grips}
Archiki Prasad, Peter Hase, Xiang Zhou, and Mohit Bansal. 2022.
\newblock Grips: Gradient-free, edit-based instruction search for prompting large language models.
\newblock \emph{arXiv preprint arXiv:2203.07281}.

\bibitem[{Pryzant et~al.(2023)Pryzant, Iter, Li, Lee, Zhu, and Zeng}]{pryzant2023automatic}
Reid Pryzant, Dan Iter, Jerry Li, Yin Lee, Chenguang Zhu, and Michael Zeng. 2023.
\newblock \href {https://doi.org/10.18653/v1/2023.emnlp-main.494} {Automatic prompt optimization with {\textquotedblleft}gradient descent{\textquotedblright} and beam search}.
\newblock In \emph{Proceedings of the 2023 Conference on Empirical Methods in Natural Language Processing}, pages 7957--7968, Singapore. Association for Computational Linguistics.

\bibitem[{Qian et~al.(2023)Qian, Han, Fung, Qin, Liu, and Ji}]{qian2023creator}
Cheng Qian, Chi Han, Yi~R Fung, Yujia Qin, Zhiyuan Liu, and Heng Ji. 2023.
\newblock Creator: Tool creation for disentangling abstract and concrete reasoning of large language models.
\newblock \emph{arXiv preprint arXiv:2305.14318}.

\bibitem[{Ravaut et~al.(2024)Ravaut, Sun, Chen, and Joty}]{ravaut2024context}
Mathieu Ravaut, Aixin Sun, Nancy Chen, and Shafiq Joty. 2024.
\newblock \href {https://doi.org/10.18653/v1/2024.acl-long.153} {On context utilization in summarization with large language models}.
\newblock In \emph{Proceedings of the 62nd Annual Meeting of the Association for Computational Linguistics (Volume 1: Long Papers)}, pages 2764--2781, Bangkok, Thailand. Association for Computational Linguistics.

\bibitem[{Shinn et~al.(2024)Shinn, Cassano, Gopinath, Narasimhan, and Yao}]{shinn2024reflexion}
Noah Shinn, Federico Cassano, Ashwin Gopinath, Karthik Narasimhan, and Shunyu Yao. 2024.
\newblock Reflexion: Language agents with verbal reinforcement learning.
\newblock \emph{Advances in Neural Information Processing Systems}, 36.

\bibitem[{Shridhar et~al.(2020)Shridhar, Yuan, C{\^o}t{\'e}, Bisk, Trischler, and Hausknecht}]{shridhar2020alfworld}
Mohit Shridhar, Xingdi Yuan, Marc-Alexandre C{\^o}t{\'e}, Yonatan Bisk, Adam Trischler, and Matthew Hausknecht. 2020.
\newblock Alfworld: Aligning text and embodied environments for interactive learning.
\newblock \emph{arXiv preprint arXiv:2010.03768}.

\bibitem[{Song et~al.(2024{\natexlab{a}})Song, Liu, Zhang, Zhang, Luo, Wang, Wu, and Wang}]{song2024adaptive}
Linxin Song, Jiale Liu, Jieyu Zhang, Shaokun Zhang, Ao~Luo, Shijian Wang, Qingyun Wu, and Chi Wang. 2024{\natexlab{a}}.
\newblock Adaptive in-conversation team building for language model agents.
\newblock \emph{arXiv preprint arXiv:2405.19425}.

\bibitem[{Song et~al.(2024{\natexlab{b}})Song, Zheng, and Luo}]{song2024counting}
Mingyang Song, Mao Zheng, and Xuan Luo. 2024{\natexlab{b}}.
\newblock Counting-stars: A simple, efficient, and reasonable strategy for evaluating long-context large language models.
\newblock \emph{arXiv preprint arXiv:2403.11802}.

\bibitem[{Su et~al.(2022)Su, Kasai, Wu, Shi, Wang, Xin, Zhang, Ostendorf, Zettlemoyer, Smith et~al.}]{su2022selective}
Hongjin Su, Jungo Kasai, Chen~Henry Wu, Weijia Shi, Tianlu Wang, Jiayi Xin, Rui Zhang, Mari Ostendorf, Luke Zettlemoyer, Noah~A Smith, et~al. 2022.
\newblock Selective annotation makes language models better few-shot learners.
\newblock \emph{arXiv preprint arXiv:2209.01975}.

\bibitem[{Wang et~al.(2025)Wang, Xu, Wang, Zhang, Yan, Zhang, Huang, and Ji}]{wang2025mobile}
Zhenhailong Wang, Haiyang Xu, Junyang Wang, Xi~Zhang, Ming Yan, Ji~Zhang, Fei Huang, and Heng Ji. 2025.
\newblock Mobile-agent-e: Self-evolving mobile assistant for complex tasks.
\newblock \emph{arXiv preprint arXiv:2501.11733}.

\bibitem[{Wang et~al.(2024{\natexlab{a}})Wang, Fried, and Neubig}]{wang2024trove}
Zhiruo Wang, Daniel Fried, and Graham Neubig. 2024{\natexlab{a}}.
\newblock Trove: Inducing verifiable and efficient toolboxes for solving programmatic tasks.
\newblock \emph{arXiv preprint arXiv:2401.12869}.

\bibitem[{Wang et~al.(2024{\natexlab{b}})Wang, Mao, Fried, and Neubig}]{wang2024agent}
Zora~Zhiruo Wang, Jiayuan Mao, Daniel Fried, and Graham Neubig. 2024{\natexlab{b}}.
\newblock Agent workflow memory.
\newblock \emph{arXiv preprint arXiv:2409.07429}.

\bibitem[{Wei et~al.(2022)Wei, Wang, Schuurmans, Bosma, Xia, Chi, Le, Zhou et~al.}]{wei2022chain}
Jason Wei, Xuezhi Wang, Dale Schuurmans, Maarten Bosma, Fei Xia, Ed~Chi, Quoc~V Le, Denny Zhou, et~al. 2022.
\newblock Chain-of-thought prompting elicits reasoning in large language models.
\newblock \emph{Advances in neural information processing systems}, 35:24824--24837.

\bibitem[{Wen et~al.(2024)Wen, Jain, Kirchenbauer, Goldblum, Geiping, and Goldstein}]{wen2024hard}
Yuxin Wen, Neel Jain, John Kirchenbauer, Micah Goldblum, Jonas Geiping, and Tom Goldstein. 2024.
\newblock Hard prompts made easy: Gradient-based discrete optimization for prompt tuning and discovery.
\newblock \emph{Advances in Neural Information Processing Systems}, 36.

\bibitem[{Wu et~al.(2024{\natexlab{a}})Wu, Wang, Yu, Zhang, Chang, and Yu}]{wu2024longmemeval}
Di~Wu, Hongwei Wang, Wenhao Yu, Yuwei Zhang, Kai-Wei Chang, and Dong Yu. 2024{\natexlab{a}}.
\newblock Longmemeval: Benchmarking chat assistants on long-term interactive memory.
\newblock \emph{arXiv preprint arXiv:2410.10813}.

\bibitem[{Wu et~al.(2024{\natexlab{b}})Wu, Bansal, Zhang, Wu, Li, Zhu, Jiang, Zhang, Zhang, Liu, Awadallah, White, Burger, and Wang}]{wu2024autogen}
Qingyun Wu, Gagan Bansal, Jieyu Zhang, Yiran Wu, Beibin Li, Erkang Zhu, Li~Jiang, Xiaoyun Zhang, Shaokun Zhang, Jiale Liu, Ahmed~Hassan Awadallah, Ryen~W White, Doug Burger, and Chi Wang. 2024{\natexlab{b}}.
\newblock \href {https://openreview.net/forum?id=BAakY1hNKS} {Autogen: Enabling next-gen {LLM} applications via multi-agent conversations}.
\newblock In \emph{First Conference on Language Modeling}.

\bibitem[{Wu et~al.(2024{\natexlab{c}})Wu, Zhao, Huang, Huang, Yasunaga, Cao, Ioannidis, Subbian, Leskovec, and Zou}]{wu2024avatar}
Shirley Wu, Shiyu Zhao, Qian Huang, Kexin Huang, Michihiro Yasunaga, Kaidi Cao, Vassilis~N Ioannidis, Karthik Subbian, Jure Leskovec, and James Zou. 2024{\natexlab{c}}.
\newblock Avatar: Optimizing llm agents for tool-assisted knowledge retrieval.
\newblock \emph{arXiv preprint arXiv:2406.11200}.

\bibitem[{Wu et~al.(2024{\natexlab{d}})Wu, Yue, Zhang, Wang, and Wu}]{wu2024stateflow}
Yiran Wu, Tianwei Yue, Shaokun Zhang, Chi Wang, and Qingyun Wu. 2024{\natexlab{d}}.
\newblock Stateflow: Enhancing llm task-solving through state-driven workflows.
\newblock \emph{arXiv preprint arXiv:2403.11322}.

\bibitem[{Wu et~al.(2022)Wu, Wang, Ye, and Kong}]{wu2022self}
Zhiyong Wu, Yaoxiang Wang, Jiacheng Ye, and Lingpeng Kong. 2022.
\newblock Self-adaptive in-context learning: An information compression perspective for in-context example selection and ordering.
\newblock \emph{arXiv preprint arXiv:2212.10375}.

\bibitem[{Xia and Luo(2025)}]{xia2025gui}
Xiaobo Xia and Run Luo. 2025.
\newblock Gui-r1: A generalist r1-style vision-language action model for gui agents.
\newblock \emph{arXiv preprint arXiv:2504.10458}.

\bibitem[{Xie et~al.(2025)Xie, Zhang, Chen, Li, Zhao, Cao, Toh, Cheng, Shin, Lei et~al.}]{xie2025osworld}
Tianbao Xie, Danyang Zhang, Jixuan Chen, Xiaochuan Li, Siheng Zhao, Ruisheng Cao, Jing~Hua Toh, Zhoujun Cheng, Dongchan Shin, Fangyu Lei, et~al. 2025.
\newblock Osworld: Benchmarking multimodal agents for open-ended tasks in real computer environments.
\newblock \emph{Advances in Neural Information Processing Systems}, 37:52040--52094.

\bibitem[{Yang et~al.(2024)Yang, Wang, Lu, Liu, Le, Zhou, and Chen}]{yang2023large}
Chengrun Yang, Xuezhi Wang, Yifeng Lu, Hanxiao Liu, Quoc~V Le, Denny Zhou, and Xinyun Chen. 2024.
\newblock \href {https://openreview.net/forum?id=Bb4VGOWELI} {Large language models as optimizers}.
\newblock In \emph{The Twelfth International Conference on Learning Representations}.

\bibitem[{Yao et~al.(2024)Yao, Yu, Zhao, Shafran, Griffiths, Cao, and Narasimhan}]{yao2024tree}
Shunyu Yao, Dian Yu, Jeffrey Zhao, Izhak Shafran, Tom Griffiths, Yuan Cao, and Karthik Narasimhan. 2024.
\newblock Tree of thoughts: Deliberate problem solving with large language models.
\newblock \emph{Advances in Neural Information Processing Systems}, 36.

\bibitem[{Yao et~al.(2023)Yao, Zhao, Yu, Du, Shafran, Narasimhan, and Cao}]{yao2023react}
Shunyu Yao, Jeffrey Zhao, Dian Yu, Nan Du, Izhak Shafran, Karthik Narasimhan, and Yuan Cao. 2023.
\newblock {ReAct}: Synergizing reasoning and acting in language models.
\newblock In \emph{International Conference on Learning Representations (ICLR)}.

\bibitem[{Yin et~al.(2024)Yin, Wang, Pan, Wan, and Wang}]{yin2024g}
Xunjian Yin, Xinyi Wang, Liangming Pan, Xiaojun Wan, and William~Yang Wang. 2024.
\newblock G$\backslash$" odel agent: A self-referential agent framework for recursive self-improvement.
\newblock \emph{arXiv preprint arXiv:2410.04444}.

\bibitem[{Yuan et~al.(2023)Yuan, Chen, Wang, Fung, Peng, and Ji}]{yuan2023craft}
Lifan Yuan, Yangyi Chen, Xingyao Wang, Yi~R Fung, Hao Peng, and Heng Ji. 2023.
\newblock Craft: Customizing llms by creating and retrieving from specialized toolsets.
\newblock \emph{arXiv preprint arXiv:2309.17428}.

\bibitem[{Zelikman et~al.(2023)Zelikman, Lorch, Mackey, and Kalai}]{zelikman2023self}
Eric Zelikman, Eliana Lorch, Lester Mackey, and Adam~Tauman Kalai. 2023.
\newblock Self-taught optimizer (stop): Recursively self-improving code generation.
\newblock \emph{arXiv preprint arXiv:2310.02304}.

\bibitem[{Zeng et~al.(2024)Zeng, Liu, Lu, Wang, Liu, Dong, and Tang}]{zeng2024agenttuning}
Aohan Zeng, Mingdao Liu, Rui Lu, Bowen Wang, Xiao Liu, Yuxiao Dong, and Jie Tang. 2024.
\newblock \href {https://doi.org/10.18653/v1/2024.findings-acl.181} {{A}gent{T}uning: Enabling generalized agent abilities for {LLM}s}.
\newblock In \emph{Findings of the Association for Computational Linguistics: ACL 2024}, pages 3053--3077, Bangkok, Thailand. Association for Computational Linguistics.

\bibitem[{Zhang et~al.(2024{\natexlab{a}})Zhang, Xiang, Yu, Teng, Chen, Chen, Zhuge, Cheng, Hong, Wang et~al.}]{zhang2024aflow}
Jiayi Zhang, Jinyu Xiang, Zhaoyang Yu, Fengwei Teng, Xionghui Chen, Jiaqi Chen, Mingchen Zhuge, Xin Cheng, Sirui Hong, Jinlin Wang, et~al. 2024{\natexlab{a}}.
\newblock Aflow: Automating agentic workflow generation.
\newblock \emph{arXiv preprint arXiv:2410.10762}.

\bibitem[{Zhang et~al.(2023)Zhang, Xia, Wang, Chen, Liu, Wu, and Liu}]{zhang2023ideal}
Shaokun Zhang, Xiaobo Xia, Zhaoqing Wang, Ling-Hao Chen, Jiale Liu, Qingyun Wu, and Tongliang Liu. 2023.
\newblock Ideal: Influence-driven selective annotations empower in-context learners in large language models.
\newblock \emph{arXiv preprint arXiv:2310.10873}.

\bibitem[{Zhang et~al.(2025)Zhang, Yin, Zhang, Liu, Han, Zhang, Li, Wang, Wang, Chen et~al.}]{zhang2025agent}
Shaokun Zhang, Ming Yin, Jieyu Zhang, Jiale Liu, Zhiguang Han, Jingyang Zhang, Beibin Li, Chi Wang, Huazheng Wang, Yiran Chen, et~al. 2025.
\newblock Which agent causes task failures and when? on automated failure attribution of llm multi-agent systems.
\newblock \emph{arXiv preprint arXiv:2505.00212}.

\bibitem[{Zhang et~al.(2024{\natexlab{b}})Zhang, Zhang, Ding, Garcia, Mallick, Madrigal, Xia, R{\"u}hle, Wu, and Wang}]{zhang2024ecoact}
Shaokun Zhang, Jieyu Zhang, Dujian Ding, Mirian~Hipolito Garcia, Ankur Mallick, Daniel Madrigal, Menglin Xia, Victor R{\"u}hle, Qingyun Wu, and Chi Wang. 2024{\natexlab{b}}.
\newblock Ecoact: Economic agent determines when to register what action.
\newblock \emph{arXiv preprint arXiv:2411.01643}.

\bibitem[{Zhang et~al.(2024{\natexlab{c}})Zhang, Zhang, Liu, Song, Wang, Krishna, and Wu}]{zhangoffline}
Shaokun Zhang, Jieyu Zhang, Jiale Liu, Linxin Song, Chi Wang, Ranjay Krishna, and Qingyun Wu. 2024{\natexlab{c}}.
\newblock Offline training of language model agents with functions as learnable weights.
\newblock In \emph{Forty-first International Conference on Machine Learning}.

\bibitem[{Zhang et~al.(2024{\natexlab{d}})Zhang, Chen, Hu, Xu, Chen, Hao, Han, Thai, Wang, Liu, and Sun}]{zhang2024bench}
Xinrong Zhang, Yingfa Chen, Shengding Hu, Zihang Xu, Junhao Chen, Moo Hao, Xu~Han, Zhen Thai, Shuo Wang, Zhiyuan Liu, and Maosong Sun. 2024{\natexlab{d}}.
\newblock \href {https://doi.org/10.18653/v1/2024.acl-long.814} {$\infty${B}ench: Extending long context evaluation beyond 100{K} tokens}.
\newblock In \emph{Proceedings of the 62nd Annual Meeting of the Association for Computational Linguistics (Volume 1: Long Papers)}, pages 15262--15277, Bangkok, Thailand. Association for Computational Linguistics.

\bibitem[{Zhang et~al.(2024{\natexlab{e}})Zhang, Bo, Ma, Li, Chen, Dai, Zhu, Dong, and Wen}]{zhang2024survey}
Zeyu Zhang, Xiaohe Bo, Chen Ma, Rui Li, Xu~Chen, Quanyu Dai, Jieming Zhu, Zhenhua Dong, and Ji-Rong Wen. 2024{\natexlab{e}}.
\newblock A survey on the memory mechanism of large language model based agents.
\newblock \emph{arXiv preprint arXiv:2404.13501}.

\bibitem[{Zhao et~al.(2024)Zhao, Huang, Xu, Lin, Liu, and Huang}]{zhao2024expel}
Andrew Zhao, Daniel Huang, Quentin Xu, Matthieu Lin, Yong-Jin Liu, and Gao Huang. 2024.
\newblock Expel: Llm agents are experiential learners.
\newblock In \emph{Proceedings of the AAAI Conference on Artificial Intelligence}, volume~38, pages 19632--19642.

\bibitem[{Zhao et~al.(2021)Zhao, Wallace, Feng, Klein, and Singh}]{zhao21calibrate}
Zihao Zhao, Eric Wallace, Shi Feng, Dan Klein, and Sameer Singh. 2021.
\newblock \href {https://proceedings.mlr.press/v139/zhao21c.html} {Calibrate before use: Improving few-shot performance of language models}.
\newblock In \emph{Proceedings of the 38th International Conference on Machine Learning}, volume 139 of \emph{Proceedings of Machine Learning Research}, pages 12697--12706. PMLR.

\bibitem[{Zhou et~al.(2022)Zhou, Muresanu, Han, Paster, Pitis, Chan, and Ba}]{zhou2022large}
Yongchao Zhou, Andrei~Ioan Muresanu, Ziwen Han, Keiran Paster, Silviu Pitis, Harris Chan, and Jimmy Ba. 2022.
\newblock Large language models are human-level prompt engineers.
\newblock \emph{arXiv preprint arXiv:2211.01910}.

\bibitem[{Zhuge et~al.(2024)Zhuge, Wang, Kirsch, Faccio, Khizbullin, and Schmidhuber}]{zhuge2024language}
Mingchen Zhuge, Wenyi Wang, Louis Kirsch, Francesco Faccio, Dmitrii Khizbullin, and Jurgen Schmidhuber. 2024.
\newblock Language agents as optimizable graphs.
\newblock \emph{arXiv preprint arXiv:2402.16823}.

\end{thebibliography}

\newpage
\appendix
\section{More Experiments and Analysis}

\begin{figure*}[t!]
	\centering
	\begin{subfigure}{0.32\linewidth}
		\centering
		\includegraphics[width=\linewidth]{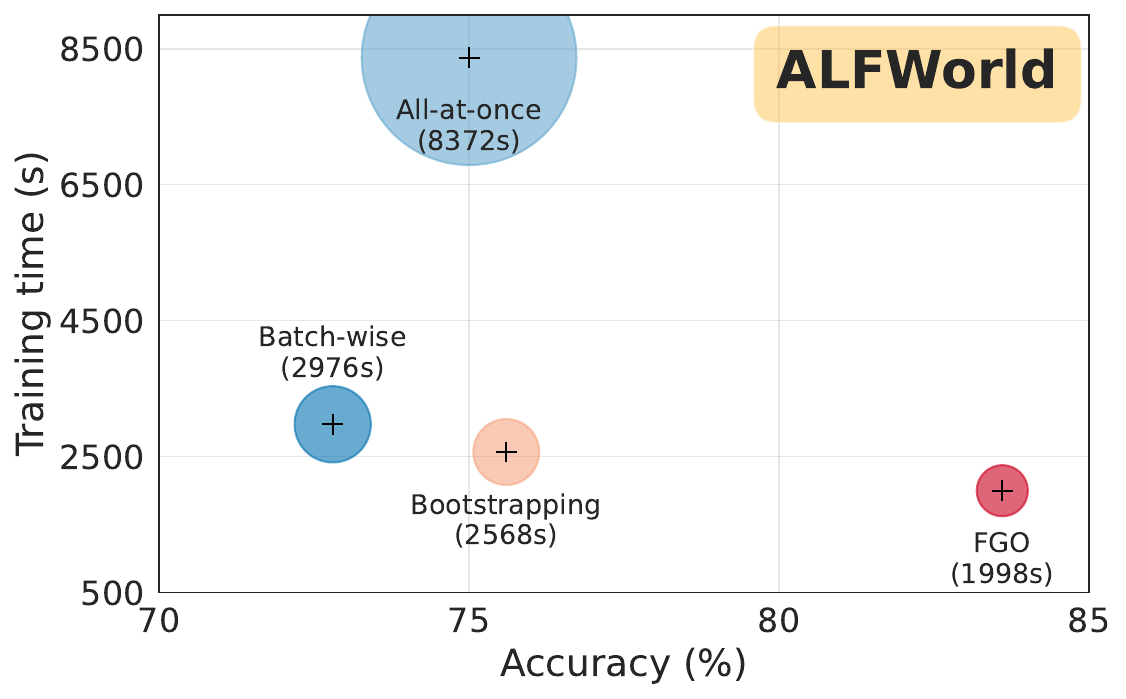}
	\end{subfigure}
	\centering
	\begin{subfigure}{0.32\linewidth}
		\centering
		\includegraphics[width=\linewidth]{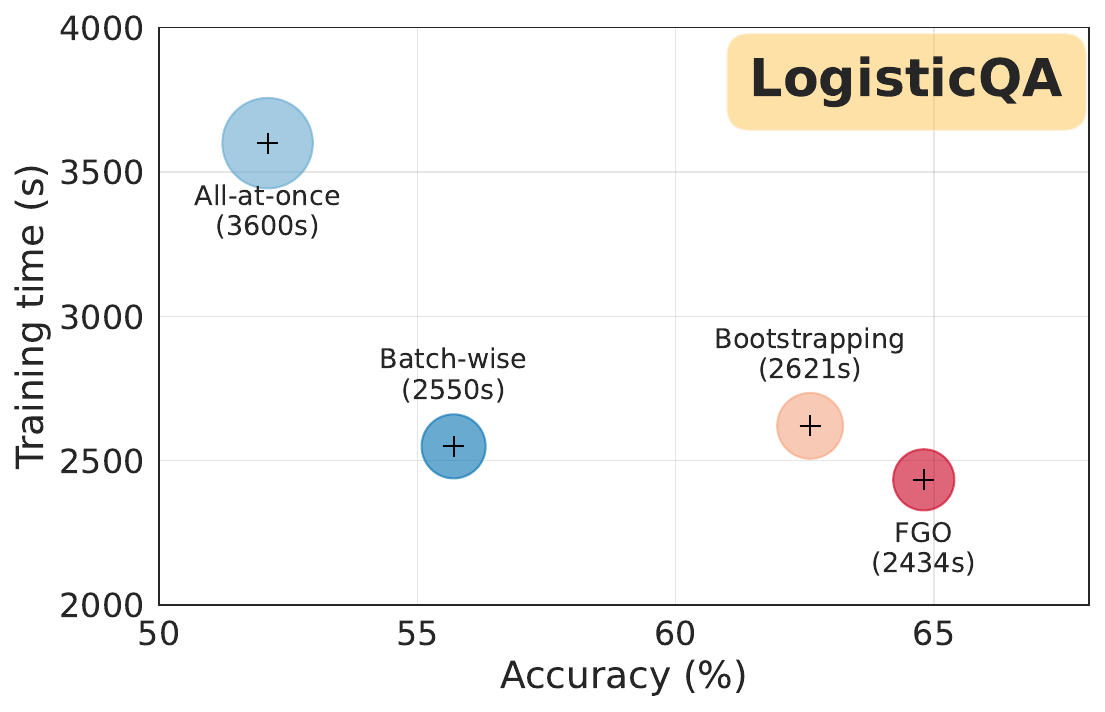}
	\end{subfigure}
	\centering
	\begin{subfigure}{0.32\linewidth}
		\centering
		\includegraphics[width=\linewidth]{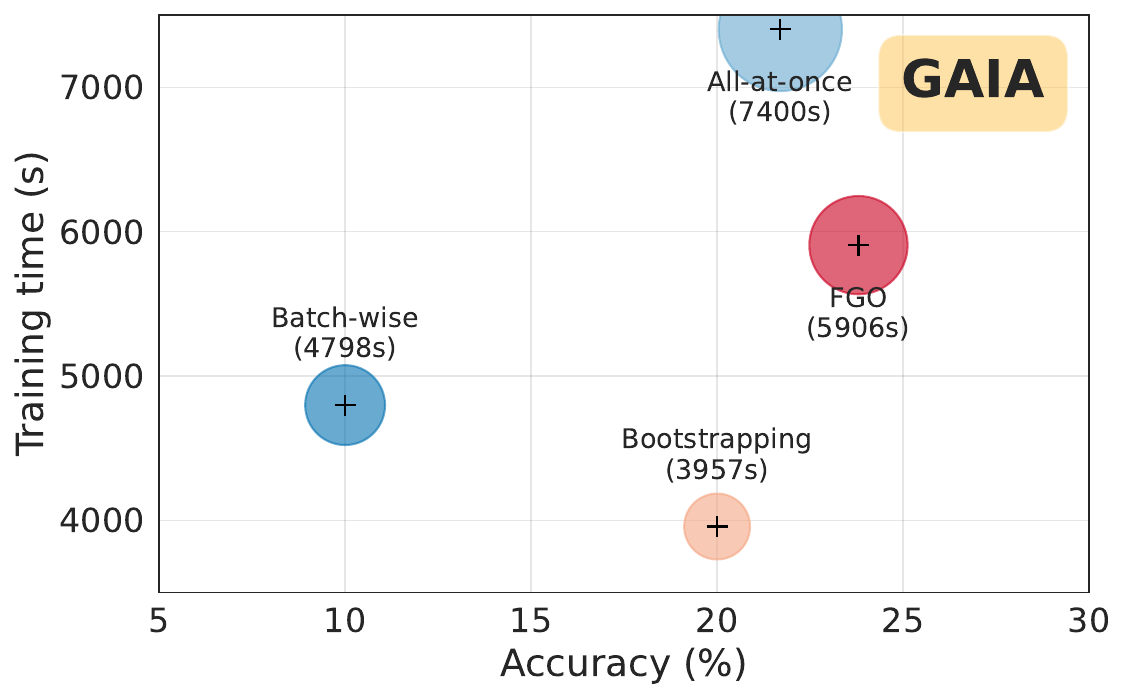}
	\end{subfigure}
	\caption{Comparison of time to optimize using different optimization methods on ALFWorld, LogisticsQA, and GAIA. Each panel plots the trained agent's performance against the time to optimize the agentic system. Circle diameters are proportional to the optimization time consumption, with crosses (+) indicating circle centers.}
    \label{fig:cost_time}
\end{figure*}

\paragraph{How is the efficiency of \ours in terms of optimizing time consumption?}
In this section, we present the time to optimize using each method. The experimental results demonstrate that \ours achieves superior optimization efficiency across all tested benchmarks. As shown in Figure \ref{fig:cost_time}, \ours consistently delivers the highest accuracy while requiring significantly less training time compared to alternatives. Notably, on ALFWorld, \ours achieves 85\% accuracy in \textasciitilde 2000 seconds, substantially outperforming All-at-once optimization by 4 times. 
The key to \ours's efficiency lies in its divide-and-conquer approach, which enables parallel optimization of dataset subsets rather than sequential processing. By distributing the computational load and then adaptively merging results, \ours effectively bypasses the scaling limitations of other optimization methods.

\paragraph{How does the trained agent perform after different optimization methods?}

\begin{figure*}[t!]
	\centering
	\begin{subfigure}{0.32\linewidth}
		\centering
		\includegraphics[width=\linewidth]{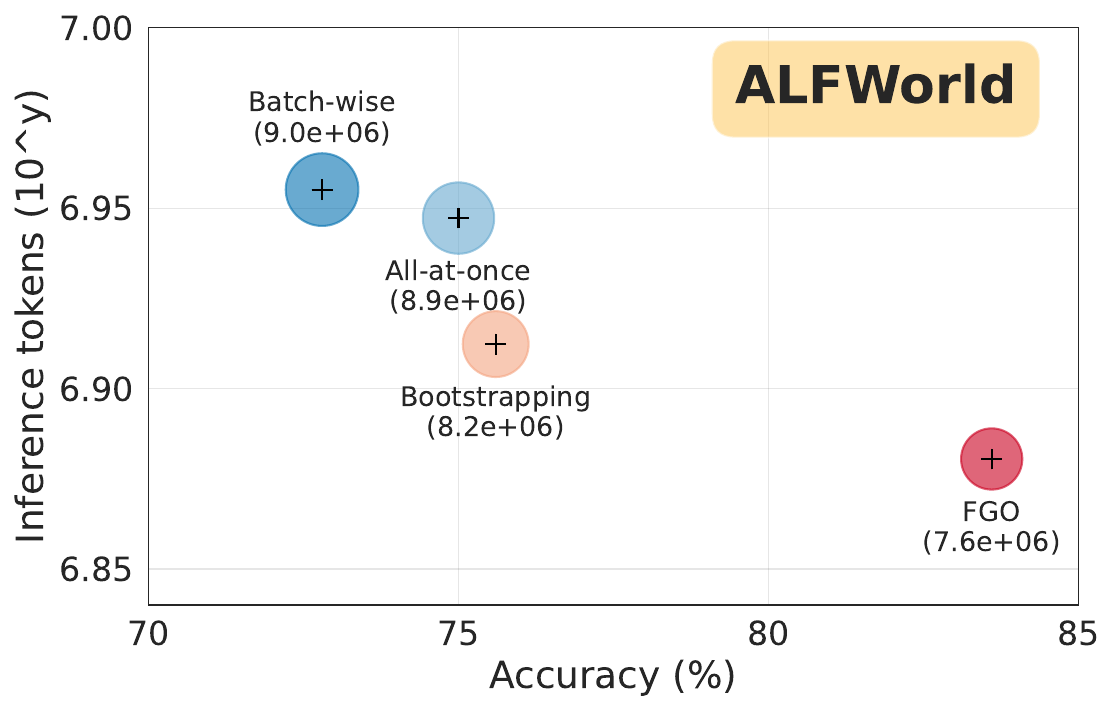}
	\end{subfigure}
	\centering
	\begin{subfigure}{0.32\linewidth}
		\centering
		\includegraphics[width=\linewidth]{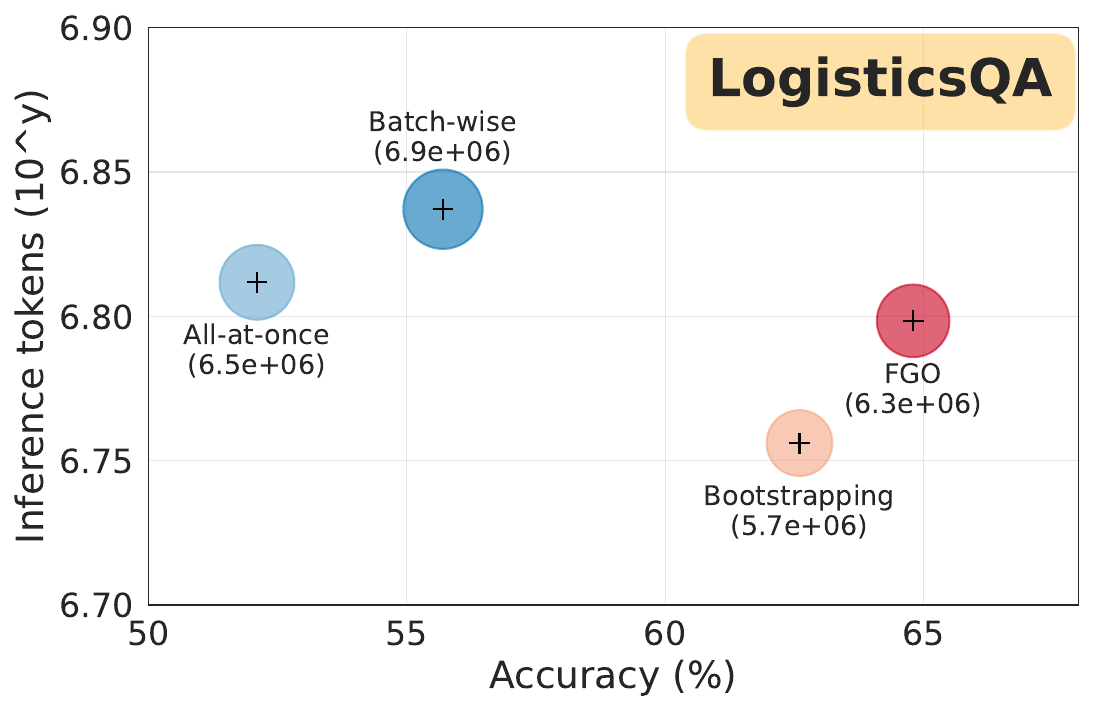}
	\end{subfigure}
	\centering
	\begin{subfigure}{0.32\linewidth}
		\centering
		\includegraphics[width=\linewidth]{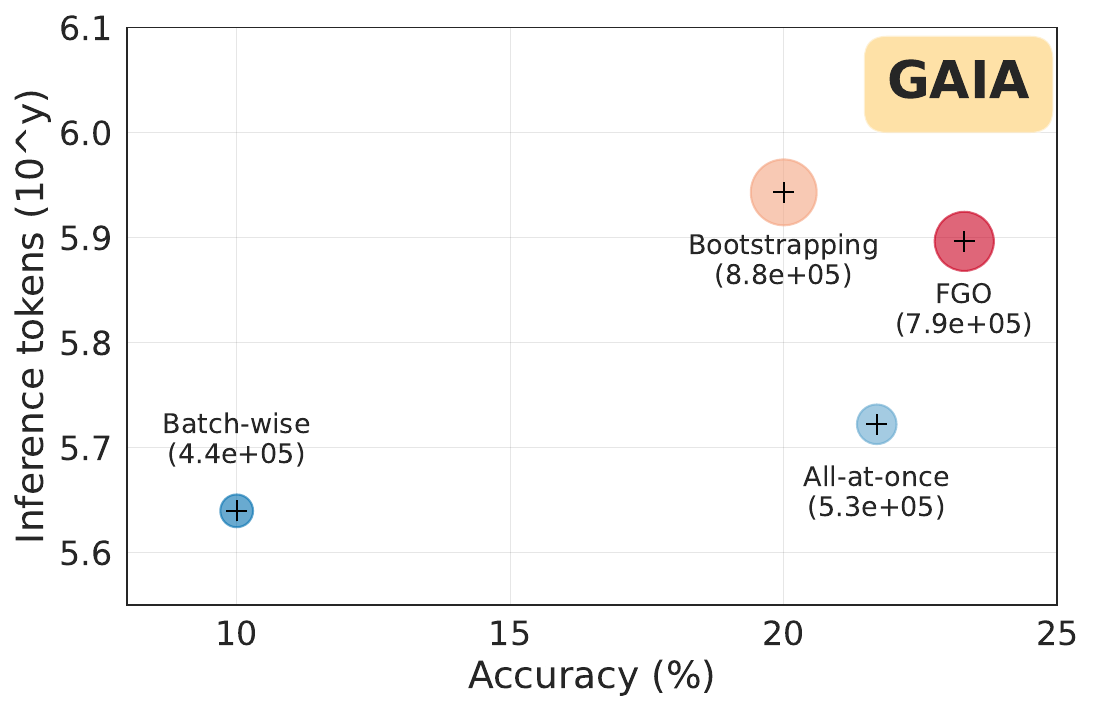}
	\end{subfigure}
	\caption{Comparison of inference token efficiency across different optimization methods on ALFWorld, LogisticsQA, and GAIA. Each panel plots the trained agent's performance against the total prompt tokens consumed during inference. Circle diameters are proportional to the inference token consumption, with crosses (+) indicating circle centers.}
    \label{fig:cost_inference}
\end{figure*}

We plot the token consumption for inferencing on the datasets with the modules trained with different methods. As shown in Figure \ref{fig:cost_inference}, agent system trained with \ours can reach the best performance with reasonable token consumption overhead.

\section{LogisticQA Dataset}
\label{anon}

\subsection{Background}

We evaluate our system on a collection of real-world Universal Business Language invoice documents, developed in cooperation with one of the world's largest logistics companies. The primary task is to extract transport reference numbers from these documents. 
The reference numbers exist in these invoice documents in a non-fixed pattern.
It typically requires human effort to extract it manually during real-world business operations.
AI agents that can effectively understand the context and extract reference numbers can make the business workflow more efficient.
The LogisticQA dataset shows LLMs' ability to achieve such a goal.
It contains 267 valid invoice documents and transport reference pairs.
It can also reflect LLM's instruction-learning capability in real-world document understanding tasks.

The dataset presents several challenging characteristics that make it an ideal testbed for evaluating the instruction learning capabilities. First, it requires specialized domain knowledge of business documents and terminology not commonly found in general language model training. Second, the hierarchical structure of UBL documents and the significant variability in format and identification patterns pose substantial extraction challenges. Additionally, as a novel benchmark without prior literature coverage, this dataset offers unique opportunities to assess agents' adaptive learning abilities in a practical, high-stakes business context.

\subsection{Dataset Statistics}

\begin{figure*}[ht]
    \centering
    \includegraphics[width=0.49\linewidth]{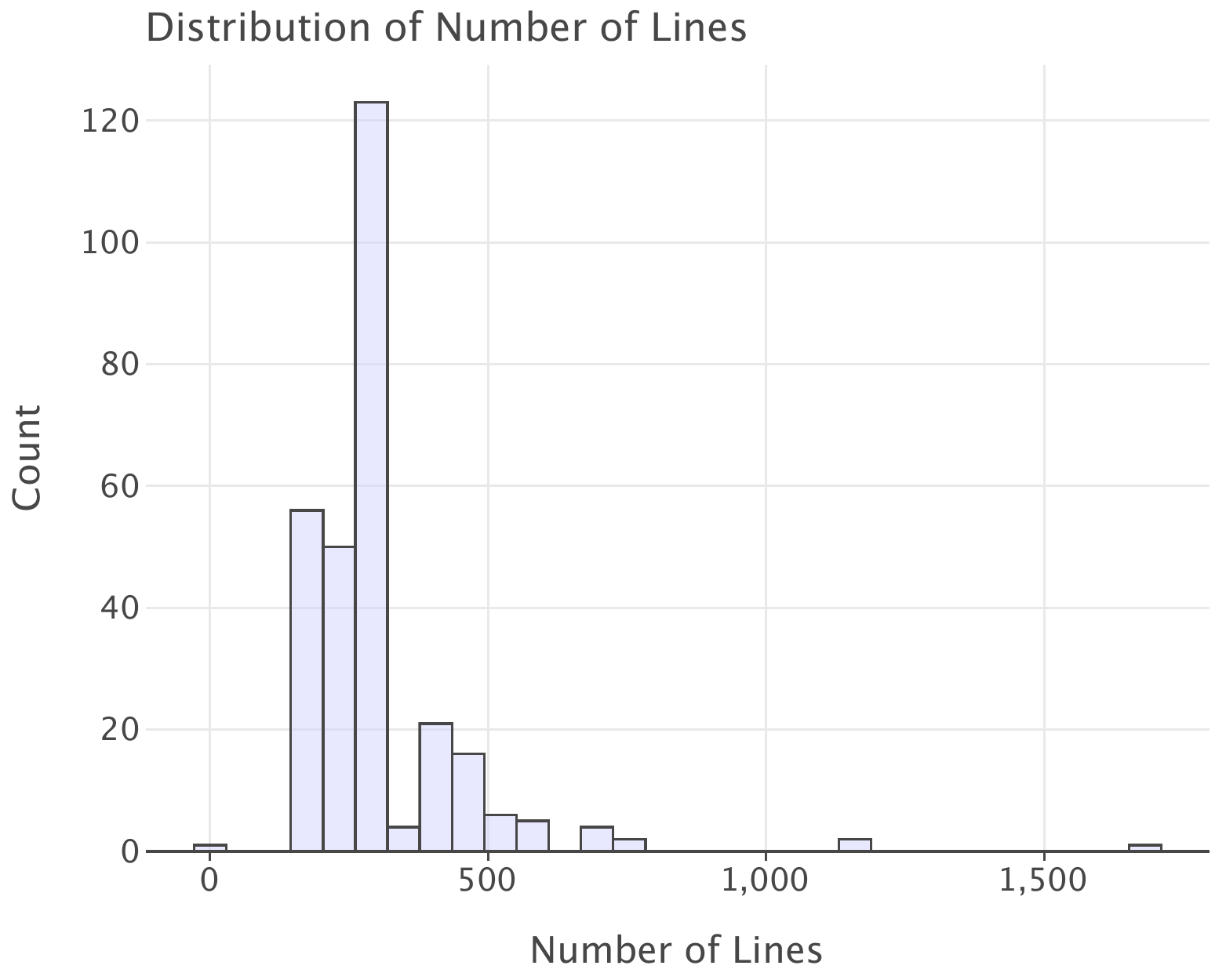}
    \includegraphics[width=0.49\linewidth]{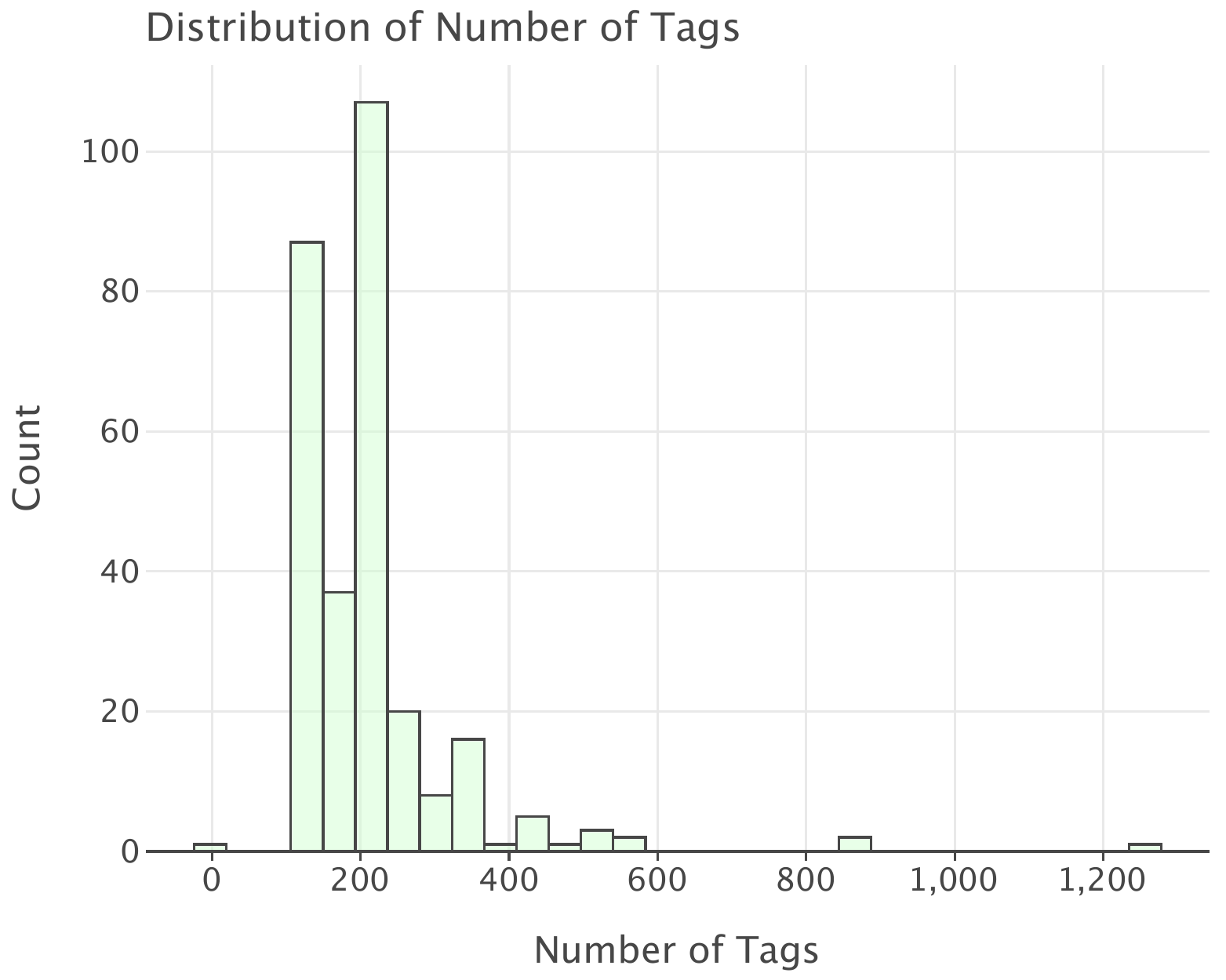}
    \includegraphics[width=0.49\linewidth]{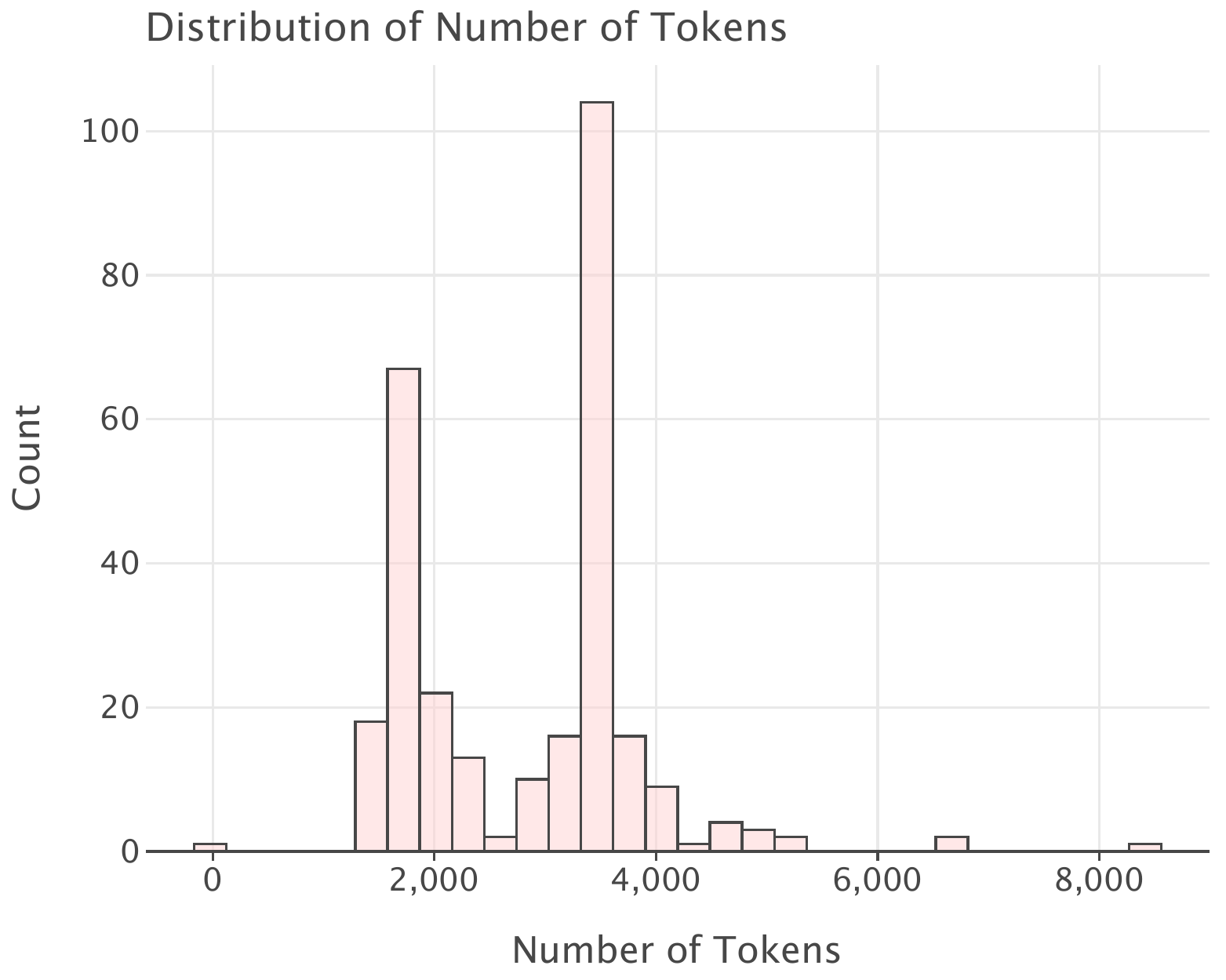}
    \includegraphics[width=0.49\linewidth]{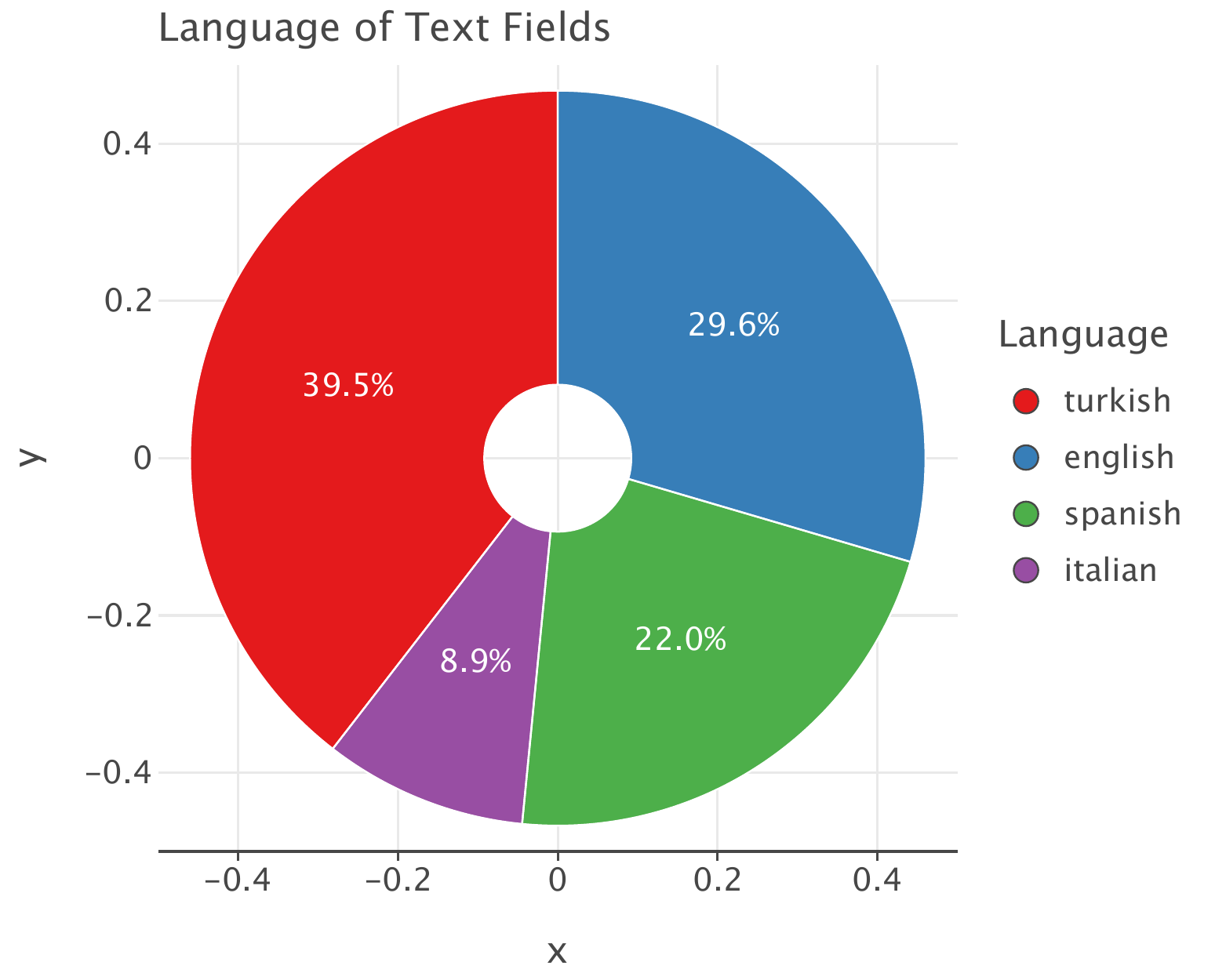}
    \caption{Statistical analysis of XML business documents. Top left: Distribution of document lengths showing typical business document sizes. Top right: Distribution of XML tags indicating document structure complexity. Bottom left: Token distribution demonstrating the long context challenge for LLM. Bottom right: Language distribution across documents reflects business documents' multinational nature.}
    \label{fig:stat-xml-doc}
\end{figure*}

The analysis of our XML business document dataset demonstrates strong alignment with real-world business documentation patterns, as shown in Figure \ref{fig:stat-xml-doc}.
The document length distribution peaks between 200-500 lines, while the XML structure complexity with most documents containing 100-400 tags. 
The token distribution centered around 2,000-4,000 tokens indicates a long-context understanding challenge for LLMs. 
Notably, the language distribution across documents (Turkish: 39.5\%, English: 29.6\%, Spanish: 22.0\%, Italian: 8.9\%) reflects a realistic multinational business environment, particularly common in European and Mediterranean operations where English serves as a lingua franca alongside regional languages.

\subsection{Dataset Example}

Here is an example XML business document in the dataset. The ground truth extraction is \texttt{847 5321 9084}. The named and loations in the dataset are all anonymized.

\lstset{language=XML}
{\tiny
\begin{lstlisting}[basicstyle=\scriptsize\ttfamily, backgroundcolor=\color{gray!5},  escapechar=|]

<?xml version="1.0" encoding="UTF-8"?>
<Invoice xmlns="urn:oasis:names:specification:ubl:schema:xsd:Invoice-2"
         xmlns:cac="urn:oasis:names:specification:ubl:schema:xsd:CommonAggregateComponents-2"
         xmlns:cbc="urn:oasis:names:specification:ubl:schema:xsd:CommonBasicComponents-2">
    <cbc:UBLVersionID>2.1</cbc:UBLVersionID>
    <cbc:CustomizationID>urn:cen.eu:en16931:2017#compliant#urn:fdc:peppol.eu:2017:poacc:billing:3.0</cbc:CustomizationID>
    <cbc:ID>rmCMsB6Km6J4Qp2a</cbc:ID>
    <cbc:IssueDate>2023-10-11</cbc:IssueDate>
    <cbc:InvoiceTypeCode>Invoice</cbc:InvoiceTypeCode>
    <cbc:DocumentCurrencyCode>TRY</cbc:DocumentCurrencyCode>
    
    <cbc:Note>SALE
HADIMKOY BRANCH 847 5321 9084
No withholding tax applies when not self-owned according to law
This invoice must be paid by: 01/08/24
PLEASE INDICATE THE VEHICLE PLATE NUMBER AND INVOICE NUMBER IN THE DESCRIPTION OF YOUR BANK TRANSFER RECEIPT
For invoices not paid by due date, late payment interest will be charged according to the Law on Collection Procedure of Public Receivables (AATUHK).
Only FourThousandThirtyTwoTL</cbc:Note>
    

    <cac:AccountingSupplierParty>
        <cac:Party>
            <cac:PartyName>
                <cbc:Name>S.S 350 COOPERATIVE AIRPORT CARGO TERMINAL LOGISTICS SERVICES MOTOR CARRIERS</cbc:Name>
            </cac:PartyName>
            <cac:PostalAddress>
                <cbc:StreetName>Cargo Terminal Cooperative Service</cbc:StreetName>
                <cbc:CityName>Springfield</cbc:CityName>
                <cbc:PostalZone>None</cbc:PostalZone>
                <cac:Country>
                    <cbc:IdentificationCode>TR</cbc:IdentificationCode>
                </cac:Country>
            </cac:PostalAddress>
        </cac:Party>
    </cac:AccountingSupplierParty>

    <cac:AccountingCustomerParty>
        <cac:Party>
            <cac:PartyName>
                <cbc:Name>GLOBAL LOGISTICS SOLUTIONS LTD.</cbc:Name>
            </cac:PartyName>
            <cac:PostalAddress>
                <cbc:StreetName>INDUSTRIAL DISTRICT SPRINGFIELD</cbc:StreetName>
                <cbc:CityName>None</cbc:CityName>
                <cbc:PostalZone>None</cbc:PostalZone>
                <cac:Country>
                    <cbc:IdentificationCode>TR</cbc:IdentificationCode>
                </cac:Country>
            </cac:PostalAddress>
        </cac:Party>
    </cac:AccountingCustomerParty>

    
    <cac:PaymentTerms>
        <cbc:Note>SALE
HADIMKOY BRANCH 847 5321 9084
No withholding tax applies when not self-owned according to law
This invoice must be paid by: 01/08/24
PLEASE INDICATE THE VEHICLE PLATE NUMBER AND INVOICE NUMBER IN THE DESCRIPTION OF YOUR BANK TRANSFER RECEIPT
For invoices not paid by due date, late payment interest will be charged according to the Law on Collection Procedure of Public Receivables (AATUHK).
Only FourThousandThirtyTwoTL</cbc:Note>
    </cac:PaymentTerms>
    

    <cac:LegalMonetaryTotal>
        <cbc:LineExtensionAmount currencyID="TRY">
            2243.26
        </cbc:LineExtensionAmount>
        <cbc:TaxExclusiveAmount currencyID="TRY">
            448.65
        </cbc:TaxExclusiveAmount>
        <cbc:TaxInclusiveAmount currencyID="TRY">
            2691.91
        </cbc:TaxInclusiveAmount>
        <cbc:PayableAmount currencyID="TRY">
            2691.91
        </cbc:PayableAmount>
    </cac:LegalMonetaryTotal>

    
    <cac:InvoiceLine>
        <cbc:ID>1</cbc:ID>
        <cbc:InvoicedQuantity unitCode="EA">1.0</cbc:InvoicedQuantity>
        <cbc:LineExtensionAmount currencyID="TRY">
            2243.26
        </cbc:LineExtensionAmount>
        <cac:Item>
            <cbc:Description>THY-NEWTOWN transportation fee-78XYZ432</cbc:Description>
            <cbc:Name>THY-NEWTOWN transportation fee-78XYZ432</cbc:Name>
        </cac:Item>
        <cac:Price>
            <cbc:PriceAmount currencyID="TRY">2243.26</cbc:PriceAmount>
        </cac:Price>
    </cac:InvoiceLine>
    
</Invoice>
\end{lstlisting}
}

\section{Complexity Analysis}

In this section, we analyze the computational complexity of the recursive clustering in the progressive merging process. 

\subsection{Clustering Tree Depth}

At each recursive step, the number of module is reduced by taking the square root:
\begin{equation}
    n_{i+1} = \sqrt{n_i}, \quad \text{with } n_0 = N.
\end{equation}
The recursion stops when the number of items satisfies:
\begin{equation}
    n_D = N^{(1/2)^D} \leq t.
\end{equation}
Taking logarithms on both sides gives:
\begin{equation}
    (1/2)^D \cdot \log N \leq \log t.
\end{equation}
Solving for \( D \) yields:
\begin{equation}
    D = O\left(\log \log N\right).
\end{equation}

\subsection{Backtesting Complexity}

Each merge operation performs a backward testing over all tasks contributing to the merged module. Since tasks are merged without duplication, the total number of unique tasks remains \( T \) throughout the process. As every level of the clustering tree processes \( T \) tasks and the depth of the tree is \( D = O(\log \log N) \), the overall complexity of testing is:
\begin{equation}
    O\left(T \cdot \log \log N\right).
\end{equation}

This demonstrates that the overhead introduced by backward testing is modest as $N$ scales.

\section{Prompt}
\subsection{ALFWorld}

\begin{lstlisting}[language={python}]

Perform actions and interact with a household to solve a task. At the beginning of your interactions, you will be given the detailed description of the current environment and your goal to accomplish. The environment only accept certain format of actions. Here are two examples, learn the pattern carefully.

\end{lstlisting}

\subsection{LogisticsQA}

\begin{lstlisting}[language={python}]

# Task background
Read the content of a xml file which contains a shipment invoice document in UBL format. You are tasked to understand the content and extract the transport reference number from it.
When you reach a conclusion, format your answer as "final answer: [extracted reference number]"

\end{lstlisting}
\subsection{GAIA}

\begin{lstlisting}[language={python}]

# Task
You need to solve the question below given by a user. When you are building tasks, explicitly consider where the task can benefit from web navigation capability.

# Task
{task}
"""


\end{lstlisting}

\section{Potential Risks}
We use close source API in this research, which can cause trouble on private dataset not intentioned for release, leading to potential privacy leakage if there is misconduct in API providers.

\end{document}